\pdfoutput=1 

\documentclass[preprints,article,accept,moreauthors]{Definitions/mdpi} 
\firstpage{1} 
\pubvolume{1}
\issuenum{1}
\articlenumber{0}
\pubyear{2026}
\copyrightyear{2026}
\datereceived{ } 
\daterevised{ } 
\dateaccepted{ } 
\datepublished{ } 

\usetikzlibrary{positioning,fit,backgrounds,arrows.meta}
\newlength{\widetablelength}
\isPreprints{\setlength{\widetablelength}{\textwidth}}{\setlength{\widetablelength}{\fulllength}}

\Title{Representing and Detecting Label Ambiguity in IMU-Based Exercise Evaluation}

\Author{Andreas Spilz \orcidA{}, Heiko Oppel \orcidB{} and Michael Munz *\orcidC{}}

\AuthorNames{Andreas Spilz, Heiko Oppel and Michael Munz}

\address[1]{%
AI for Sensor Data Analytics Research Group, Ulm University of Applied Sciences,  89081 Ulm, Germany}

\corres{\hangafter=1 \hangindent=1.05em \hspace{-0.82em} Correspondence: michael.munz@thu.de}

\abstract{Home-based physiotherapy is performed without supervision, which leads to incorrect execution and motivates systems that assess movement automatically from inertial measurement units (IMUs). Such systems assign each repetition to a category, yet a relevant share of repetitions fall near a class boundary, where even trained raters disagree. Classifiers trained with one-hot labels collapse these borderline repetitions onto a single class and discard this ambiguity. 
{To address this, we build on label distribution learning, which represents each repetition as a distribution over classes instead of a single label. We introduce a way to construct such distributions without a large rater pool by perturbing the thresholds of a rule-based evaluation procedure to simulate rater disagreement. We train a network to reproduce these distributions with a Kullback--Leibler objective, which we call the ambiguity approach, and compare it against a one-hot cross-entropy baseline on four IMU exercise datasets. From the predicted distribution we then determine whether a repetition is ambiguous and which classes are relevant to it.}
The ambiguity approach matched or exceeded the baseline classification on all four datasets and detected ambiguity and the relevant classes more reliably. Representing the label distribution in the training target therefore adds information about ambiguity at no cost to classification.}


\addhighlights{yes}
\renewcommand{\addhighlights}{%

\noindent\textbf{What are the main findings?}
 \begin{itemize}[labelsep=2.5mm,topsep=-3pt]
 \item Automatically generated label distributions (AGLDs) extend a rule-based labeling procedure with simulated rater variability and yield a class distribution per repetition, capturing label ambiguity without a large human rater pool.
 \item \textls[-15]{A network trained to reproduce the full label distribution matches or exceeds a one-hot cross-entropy baseline in classifying the most relevant class on four IMU exercise datasets while detecting ambiguous repetitions and their competing classes more~reliably.}
 \end{itemize}\vspace{3pt}
\textbf{What are the implications of the main findings?}
 \begin{itemize}[labelsep=2.5mm,topsep=-3pt]
 \item Representing label ambiguity in the training target adds information about borderline repetitions at no cost to classification performance.
 \item An automated feedback system can flag repetitions that lie between two evaluation categories and report both competing classes instead of forcing borderline executions onto a single class.
 \end{itemize}
}

\begin{document}

\section{Introduction}
\label{sec:introduction}

Home-based training is a central part of many physiotherapeutic treatments and improves patient outcomes~\citep{ashari_effectiveness_2016, latham_effect_2014, gelaw_effectiveness_2020, flynn_home-based_2019}. These exercises are performed without professional supervision, which reduces adherence to the prescribed regimen~\citep{argent_patient_2018} and leads to incorrect execution~\citep{faber_majority_2015}. This slows progress and can cause inappropriate loading or injury. Systems that monitor the movement and return feedback to the user address these issues and have been shown to improve adherence~\citep{lang_digital_2022}. Wearable inertial measurement units (IMUs) combined with deep learning offer one such system, an objective and scalable alternative to expert observation~\citep{swainRoleMultiSensorMeasurement2023, leeAutomaticClassificationSquat2020}. Such a system assigns each repetition to an evaluation category. Most repetitions clearly belong to one category, but a relevant share fall close to the boundary between two, where even trained raters arrive at different scores. The limited interrater reliability of established criteria-based schemes such as the functional movement screening (FMS)~\citep{cook_functional_2014, cook_functional_2014-1} bears this out since rating each repetition against a fixed set of defined criteria still leaves room for disagreement~\citep{shultz_test-retest_2013}.

This disagreement follows from how the criteria are applied. A rater judges quantities such as joint angles or the relative position of body segments visually with limited precision. In the FMS deep squat, for example, one criterion requires the femur to be below the horizontal at the lowest point, yet neither the anatomical reference points nor the definition of horizontal are precisely specified, so they differ slightly between raters. This creates a borderline region in which one rater counts the criterion as just fulfilled and another as just violated, so a repetition that falls into it cannot be assigned to a single class with confidence. We refer to this disagreement at the class boundary as ambiguity. Ambiguity of this kind is not confined to exercise evaluation. Other criteria-based clinical scores show the same variation between raters~\citep{kottner_systematic_2009}, and, where it has been examined closely, this variation concentrates on cases near a decision threshold while ratings far from it stay clear-cut~\citep{dalton_histologic_2000}. The underlying phenomenon is general, but we address it here for IMU-based movement~assessment.

Deep learning classifiers trained with one-hot labels and the associated losses cannot represent this ambiguity. They are optimized to assign the full probability mass to a single class and therefore collapse every rating of a repetition onto one category, discarding the information that a notable share of raters would have assigned a second category. We call such repetitions ``borderline repetitions''. This becomes a problem once the decision is communicated to a patient. A hard decision boundary forces two nearly identical borderline repetitions onto opposite sides as soon as each tips only marginally one way or the other, so executions that feel the same to the patient receive different verdicts. Without any indication that these repetitions were borderline, the diverging feedback appears arbitrary and undermines trust in the assessment. A model that signals that a repetition lies between two categories resolves this because it frames the differing outcomes as a genuine borderline situation rather than a contradiction.

One possible representation of this kind, a distribution over categories instead of a single label, is label distribution learning (LDL)~\citep{gengLabelDistributionLearning2016}. LDL replaces the single label with a distribution that assigns each class a description degree quantifying how strongly it applies to an instance and trains a model to reproduce this distribution rather than a single class. Gao et al. transferred this idea to deep networks with deep label distribution learning (DLDL), minimizing the Kullback--Leibler divergence (KLD) between the predicted and target distributions end to end, which improved tasks such as age and head-pose estimation and reduced overfitting on small datasets~\citep{gaoDeepLabelDistribution2017a}.

One obstacle in applying LDL is that label distributions are rarely available as most datasets provide a single label per instance and obtaining a distribution directly is costly. Each method proposed to construct such a distribution therefore relies on an additional assumption or resource. DLDL constructs the target distribution by placing a normal distribution around the single {original} label~\citep{gaoDeepLabelDistribution2017a}, which presupposes an ordered label space where neighboring labels are meaningful, as in age or pose{, and assigns the same distribution to every instance of a class}. Label enhancement instead recovers a distribution from logical labels through prior knowledge, such as neighboring-label similarity, formulated by Gao et al. as a reinforcement-learning process~\citep{gaoLabelEnhancementLabel2020}{, so the distribution is reconstructed from the label rather than derived from the process that causes the disagreement}. An alternative is the vote strategy, which aggregates the scores of many annotators into an empirical distribution and therefore requires a large and qualified rater pool. {Our construction follows the vote strategy but replaces the human raters with simulated ones that apply the evaluation criteria with slightly different thresholds (Section~\ref{sec:aglv}). This yields a distribution that is specific to the individual repetition and requires neither an ordered label space nor a rater~pool.}

A related body of work addresses difficulties that surround the labels rather than the ambiguity itself and in each case commits to a single label. Representative approaches are reviewed in the following.

Methods for learning with noisy labels treat the observed annotations as corrupted versions of an underlying true label and aim to recover that label, in some cases by relaxing the target to a set of candidate classes to reduce the effect of annotation errors~\citep{lienenMitigatingLabelNoise2024}. Multi-annotator approaches model the disagreement between raters to infer one latent ground truth~\citep{liConfusionMatrixLearning2023}, which requires several annotations per instance and treats disagreement as noise to be removed. Ranking-based networks take a different route and exploit the order of the labels to predict an ordinal score~\citep{chenUsingRankingCNNAge2017}, which assumes an ordered label space and returns a single label rather than a distribution. None of these preserves the ambiguity of a repetition, which in our setting is the quantity the model should retain and report.

The work closest to ours targets functional movement screening with label distributions directly. Lin et al.~\citep{linAutomaticEvaluationFunctional2023, linAutomaticEvaluationMethod2024, linAutomaticEvaluationMethod2024a} train deep networks to predict FMS score distributions with a KLD objective across attention-based, dual-stream, and encoder--decoder architectures. Their target distributions are generated from a normal distribution centered on the original score. This smooths the ordinal scale in the manner of age estimation, where adjacent labels contribute with decreasing probability, but it does not reflect how evaluators actually disagree on a given repetition. They report only the distributional prediction without contrasting it against a one-hot baseline, so the effect of distributional training on the classification itself remains open. Identifying which repetitions are ambiguous was not their aim, so their method does not provide the criterion a feedback system would need to flag such repetitions.

Three gaps follow from this. First, no method derives a label distribution for movement assessment that accounts for the difficulties inherent to this task, the borderline repetitions and the resulting disagreement, without resorting to a large rater pool. Second, no study compares label-distribution training against a one-hot baseline for automatic movement classification. Third, no established method determines from the predicted output distribution whether a repetition lies between two classes, so even a faithful distribution provides no agreed criterion for flagging the ambiguous repetitions it encodes.

We address these gaps and make three contributions.
First, building on our previous work on the automatic assessment of movement exercises from kinematic data~\citep{spilzBoostingAutomaticExercise2025}, we present an approach that derives a label distribution for a repetition from an artificially generated pool of raters{, which is the central new element of this work}.
Second, {following established deep label distribution learning,} we train a network to reproduce the full label distribution with a KLD objective, which we refer to as the ambiguity approach, and compare it against a one-hot cross-entropy baseline on four IMU-based exercise datasets{, a comparison that has not previously been reported in this setting}.
Third, we present an evaluation methodology that uses the predicted distribution to determine whether a repetition is ambiguous and which classes are most relevant to its assessment. {The quantities it builds on, the entropy of the distribution and its two most probable classes, are established. What is new is their use as an explicit criterion for flagging ambiguous repetitions and the systematic evaluation of this criterion in this domain.}

\section{Materials and Methods}
\label{sec:methods}

\subsection{Datasets}
\label{sec:datasets}

We evaluate the proposed approach on four datasets recorded by our group in two~earlier measurement studies~\citep{spilz_automatic_2023, spilz_gaitex_2025}. Two of them cover rehabilitation exercises used in the treatment of foot drop, the resisted dorsiflexion (RD) and the resisted gait simulation (RGS). In RD the participant lifts the foot at the ankle against resistance, and, in RGS, the participant reproduces gait phases against resistance. 
The other two cover FMS tasks~\citep{cook_functional_2014,cook_functional_2014-1}, the deep squat (DS) and the hurdle step (HS). The full measurement protocols and study designs are described in the original publications~\citep{spilz_automatic_2023, spilz_gaitex_2025}, and only the most important setup is repeated here for the sake of brevity.

The RD and RGS datasets were captured with nine Xsens MVN Awinda IMUs (Movella, El Segundo, CA, USA), reduced from the full-body scheme to the lower limbs and pelvis, with an additional toe-mounted sensor on the right foot to resolve foot kinematics. The DS and HS datasets were captured with 15 Shimmer3 IMUs (ShimmerSensing, Dublin, Ireland) following the Xsens MVN full-body configuration~\citep{roetenberg}. Both systems provide tri-axial accelerometers, gyroscopes, and magnetometers. The Xsens units sampled at 100~Hz, the Shimmer3 units at 120~Hz.

All the recordings were collected under ethically approved research protocols. Ethical approval was granted by the Ethics Committee of Ulm University of Applied Sciences (reference numbers 2021-01 and 2024-01).
A written consent form was distributed to all the participants prior to the measurements and signed by each participant, covering data acquisition, storage, and anonymized analysis for research purposes.

\begin{figure}[H]

\includegraphics[width=\textwidth]{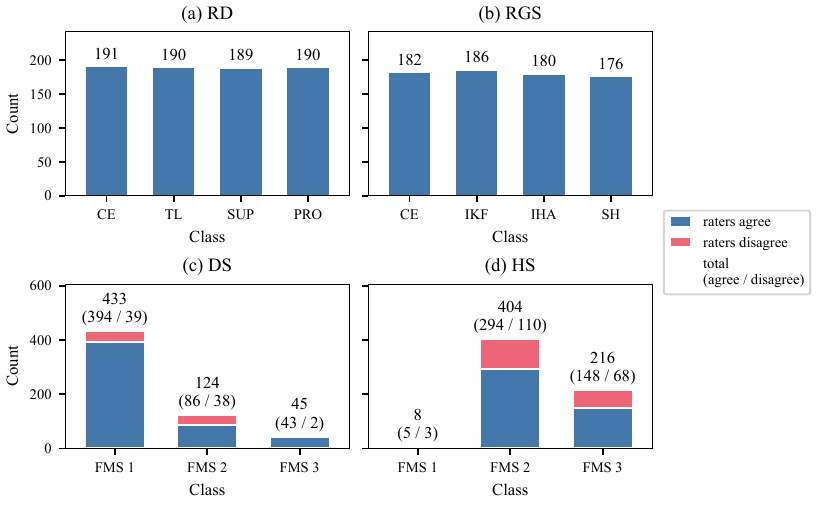}
\caption{Number of repetitions per class for each dataset. For RD the classes are CE, TL, SUP, and PRO, for RGS they are CE, IKF, IHA, and SH, and, for DS and HS, they are the FMS scores 1, 2, and 3 (Section~\ref{sec:datasets}). For DS and HS each bar is split into the count of repetitions on which the raters assigned the same score and the count on which they disagreed, given in parentheses as (agree/disagree). RD and RGS carry a single instruction-based label per repetition and are shown without a split.\label{fig:class-dist}}
\end{figure}

RD (about 760 repetitions from 19 participants) and RGS (about 720 repetitions from 18 participants) each contain four execution categories. For RD these are correct execution (CE), toe lifted (TL), supination (SUP), and pronation (PRO). For RGS they are correct execution (CE), increased knee flexion (IKF), increased hip abduction (IHA), and static hip (SH). Across both datasets the categories distinguish correct execution from qualitatively distinct movement faults rather than ordered degrees of a single deviation and therefore form a nominal scale. Both datasets are close to balanced across these four classes (Figure~\ref{fig:class-dist}).

DS (about 600 repetitions from 15 participants) and HS (about 620 repetitions from 15 participants) are scored on the three-point FMS scale. Categories 1, 2, and 3 are ordinal, with a higher score indicating a cleaner execution. Both datasets are imbalanced, with FMS~1 most frequent in DS and FMS~2 most frequent in HS, with FMS~1 occurring in only eight~repetitions.

The original labels differ in origin across the two studies. For RD and RGS the labels are instruction-based: each repetition was performed on instruction to reproduce a specified execution or fault, and the instruction defines the label~\citep{spilz_gaitex_2025}. For DS and HS the labels were assigned by three to five human raters using the FMS scale~\citep{spilz_automatic_2023}. In both datasets a notable number of repetitions received differing scores from these raters, and the count per class is shown in Figure~\ref{fig:class-dist}. Such disagreement can be read as a sign of ambiguity. However, the raters were not asked whether a repetition is ambiguous, so this signal rests on differing scores alone.

\subsection{Generation of the Automatically Generated Label Distributions (AGLDs)}
\label{sec:aglv}

The rule-based automatic labeling procedure from our augmentation work~\citep{spilzBoostingAutomaticExercise2025} is the basis for generating the label distributions introduced in this work. We briefly recap how it assigns a single label to a repetition. Each exercise is described by a set of binary evaluation criteria defined by physiotherapists, following the defined error patterns for RD and RGS and the established FMS assessment for DS and HS. A specific pattern of fulfilled and violated criteria then maps a given repetition to one class label. Every criterion is mapped to one or multiple kinematic quantities, such as a joint angle, a segment position, or a distance between anatomical points. These quantities are computed from an OpenSim~4.5
inverse-kinematics simulation of the repetition~\citep{delp_opensim_2007}. A criterion is treated as fulfilled or violated by comparing its kinematic quantity against a threshold. As the needed thresholds are often not described in the literature, we derived them from the available data. In our augmentation work~\citep{spilzBoostingAutomaticExercise2025}, we used a random search that selects the threshold set that maximizes the agreement between the automatic labels and the original labels, measured as the geometric mean of the class-wise $F_1$-scores.

The procedure assigns a single label to each repetition. Human raters trained on the same criteria, however, do not always agree{ because they effectively apply slightly different thresholds (Section~\ref{sec:introduction}).}

\begin{figure}[H]
\centering
\noindent\makebox[\textwidth][c]{%
\begin{tikzpicture}[
  node distance=10mm,
  box/.style={draw, rounded corners, align=center, inner sep=4pt, minimum height=8mm, text width=52mm, font=\small},
  data/.style={box, fill=black!5},
  op/.style={box, fill=blue!5},
  result/.style={box, fill=green!8},
  arr/.style={-{Latex[length=2mm]}, thick},
  dat/.style={font=\small\itshape, xshift=-1mm, align=right, overlay},
]
\node[data] (imu) {IMU data of repetition $n$};
\node[op, below=of imu] (ik) {Inverse kinematics (OpenSim)};
\node[op, below=of ik] (thr) {Sample thresholds\\$T_c^{(i)}\sim\mathcal{N}(\mu_c,\sigma_c^2)$};
\node[op, below=of thr] (crit) {Evaluate criteria and derive class $\hat{y}^{(i)}$};
\node[op, below=of crit] (freq) {Calculate relative class frequencies};
\node[result, below=of freq] (agld) {AGLD $\mathbf{p}^{(n)}$};
\draw[arr] (imu)--(ik);
\draw[arr] (ik)--(thr) node[dat, midway, left] {kinematic quantities};
\draw[arr] (thr)--(crit) node[dat, midway, left] {$T_c^{(i)}$};
\draw[arr] (crit)--(freq) node[dat, midway, left] {$\hat{y}^{(i)}$};
\draw[arr] (freq)--(agld);
\begin{scope}[on background layer]
\node[draw, dashed, rounded corners, fit=(thr)(crit), inner sep=6pt,
      label={[font=\small\itshape, align=left, overlay]right:{repeat\\$i=1,\dots,100$}}] {};
\end{scope}
\end{tikzpicture}%
}
\caption{Pipeline for generating the AGLD of a single repetition $n$ (Section~\ref{sec:aglv}). The 100 threshold sets $T_c^{(i)}$ defining the virtual raters are drawn once and kept fixed across all the repetitions.\label{fig:agld-pipeline}}
\end{figure}
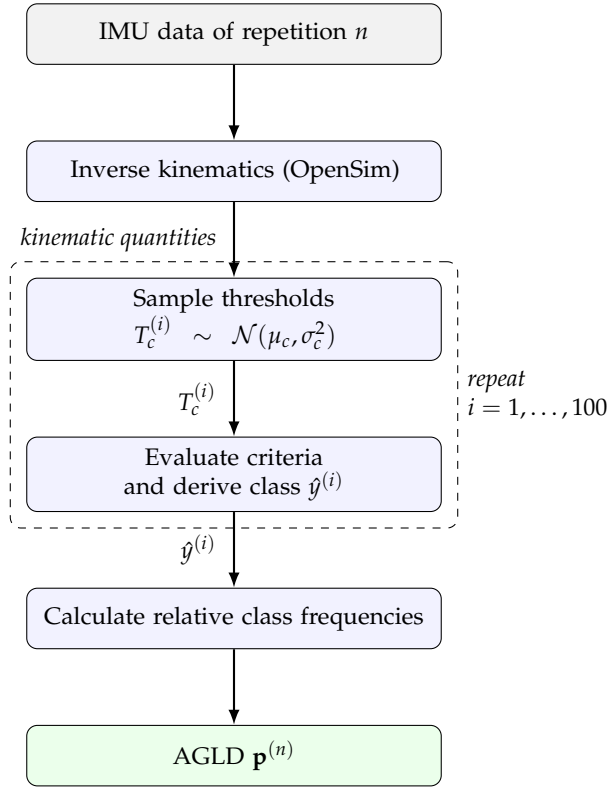

We reproduce this variability between raters with 100 instances of the described automatic labeling procedure. All the instances share the same criteria and the same ruleset, and they differ only in the thresholds applied to the individual criteria. For each instance $i$, the threshold $T^{(i)}_c$ of a criterion $c$ is drawn from a normal distribution 

\[
T_c^{(i)} \sim \mathcal{N}(\mu_c, \sigma_c^2),
\]
where $\mu_c$ is the best-fit threshold for criterion $c$ obtained from the random search described above, and $\sigma_c$ is a heuristically set standard deviation according to the measured quantity. We set $\sigma_c = 5^\circ$ for angular criteria, $\sigma_c = 5$~cm for distance criteria and $\sigma_c = 0.5~\frac{\text{m}}{\text{s}}$
for velocity criteria. Each instance assigns a single class to a repetition, and the relative frequencies of these assignments across the 100 instances define its automatically generated label distribution (AGLD). For a repetition $n$, we denote this distribution by $\mathbf{p}^{(n)} = (p^{(n)}_1, \ldots, p^{(n)}_K)$
over the $K$ classes of the respective dataset, where $p^{(n)}_k$ is the relative frequency of class $k$ across the 100 instances. {Figure~\ref{fig:agld-pipeline} summarizes this generation pipeline.}

{The generation of the AGLD is identical for all four datasets and would extend to any further exercise whose assessment rests on a ruleset that can be evaluated on the kinematic quantities of an inverse-kinematics simulation. The original labels differ in origin between the foot drop and  FMS exercises (Section~\ref{sec:datasets}), but this difference does not carry over to the AGLD because the labels do not enter the pipeline directly. They are used only to fit the means $\mu_c$ of the threshold distributions~\citep{spilzBoostingAutomaticExercise2025}. In particular, the instruction-based labels of RD and RGS do not make their repetitions unambiguous. An instruction fixes only the intended execution of a repetition, not its executed kinematics, so a repetition performed on instruction can still fall close to a criterion threshold. The labeling instances then disagree on it and its AGLD spreads over several classes, which is visible for RD and RGS in Figure~\ref{fig:entropy-dist} (Section~\ref{sec:entropy}).}

\subsection{Ambiguity and Entropy}
\label{sec:entropy}

The AGLD assigns each repetition a probability distribution over the classes rather than a single label. To determine whether one class or several classes are relevant for the evaluation of a given repetition, this distribution has to be reduced to a scalar quantity.

We use the Shannon entropy of the AGLD for this purpose. For a repetition with label distribution $\mathbf{p} = (p_1, \ldots, p_K)$ over the $K$ classes of the respective dataset, the entropy is
\[
H(\mathbf{p}) = -\sum_{k=1}^{K} p_k \log_2 p_k,
\]
measured in bits. The entropy is low when the distribution concentrates on a single class, which corresponds to an unambiguous repetition, and it is high when the distribution spreads across several classes, which corresponds to an ambiguous repetition. Entropy thereby provides a graded measure of ambiguity.

\begin{figure}[H]

\includegraphics[width=\textwidth]{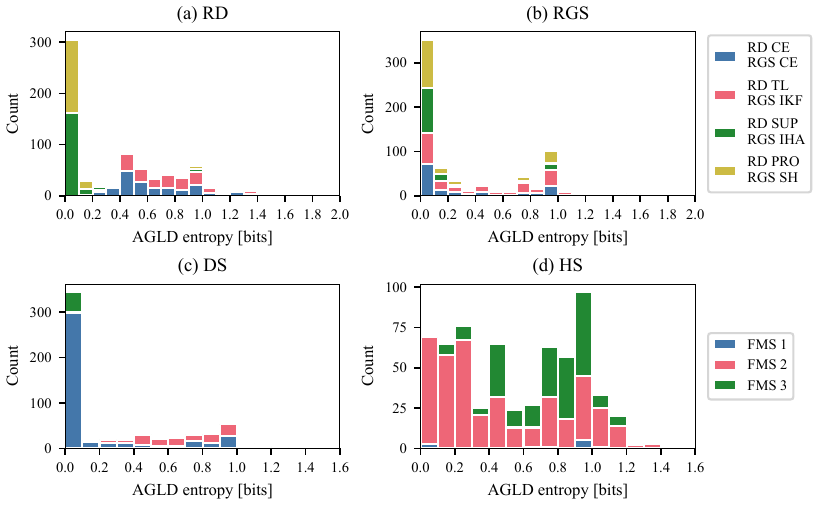}
\caption{Distribution of the AGLD entropy across repetitions, shown separately for each dataset. Entropy is given in bits. Within each bar the counts are stacked by the most frequent class of the AGLD, the class assigned the largest share across the 100 labeling instances. The per-exercise classes are defined in Section~\ref{sec:datasets}. The maximum attainable entropy differs between datasets, with $\log_2 4 = 2$~bit for the four-class datasets RD and RGS and $\log_2 3 \approx 1.58$~bit for the three-class datasets DS and HS, which sets the differing extent of the horizontal axis.
\label{fig:entropy-dist}}
\end{figure}

Figure~\ref{fig:entropy-dist} shows the distribution of the AGLD entropy for each dataset. {Entropies beyond zero occur in all four datasets, including RD and RGS, for the reason given in Section~\ref{sec:aglv}.}

\vspace{-6pt}

\subsection{Network and Experimental Design}
\label{sec:training}

We adopt the input representation and the network from our augmentation work~\citep{spilzBoostingAutomaticExercise2025} and summarize the relevant configuration here.

Each repetition is represented as a set of segment orientation sequences in quaternion form. The sequences are resampled to a fixed length of 256 time steps by spherical linear interpolation~\citep{shoemake_animating_1985} and arranged row-wise per IMU. This yields an input of shape \mbox{$(N_{\mathrm{IMU}} \times 4) \times 256$}, where $N_{\mathrm{IMU}}$ is the number of sensors and the factor four corresponds to the quaternion components of one IMU.

The classifier is a convolutional network with two convolutional blocks followed by a fully connected stage. 
The first block reduces the four quaternion components of each IMU to a per-sensor representation, the second block extracts temporal patterns along the resampled time axis, and the fully connected stage maps the resulting features to the $K$ classes of the respective dataset through a softmax output. {Batch normalization and rectified linear unit (ReLU) activations are used throughout, and a dropout rate of $0.2$ is applied in the fully connected stage. The full layer configuration is summarized in Table~\ref{tab:architecture} and follows~\citep{spilzBoostingAutomaticExercise2025}.}

We compare two training configurations that differ only in the label representation derived from the same AGLD so that any difference in the results follows from the training objective rather than from the data. We refer to them as the baseline and the ambiguity approach. For repetition $n$, let $\mathbf{p}^{(n)} = (p^{(n)}_1, \ldots, p^{(n)}_K)$ denote its AGLD and $\mathbf{q}^{(n)} = (q^{(n)}_1, \ldots, q^{(n)}_K)$ the softmax output of the network.

\begin{table}[H]
\caption{{Layer configuration of the convolutional classifier (Section~\ref{sec:training}). Kernel and stride are given as (sensor axis $\times$ time axis), and $K$ is the number of classes of the respective dataset. BN, batch normalization; FC, fully connected.}\label{tab:architecture}}

\begin{tabularx}{\textwidth}{lLccc}
\toprule
\textbf{Stage} & \textbf{Layer} & \textbf{Kernel} & \textbf{Stride} & \textbf{Channels}/\textbf{Units} \\
\midrule
Input   &                                       &         &         & $(N_{\mathrm{IMU}} \times 4)\times 256$ \\
Block 1 & Conv2D + BN + ReLU                    & $(4,1)$ & $(4,1)$ & 32 \\
Block 2 & Conv2D + BN + ReLU                    & $(1,8)$ & $(1,6)$ & 64 \\
        & Flatten                               &         &         &  \\
FC      & Linear + ReLU + Dropout ($0.2$)       &         &         & 64 \\
Output  & Linear + Softmax                      &         &         & $K$ \\
\bottomrule
\end{tabularx}
\end{table}

The baseline treats each repetition as unambiguous. It collapses the AGLD to a one-hot target $\mathbf{y}^{(n)}$ by assigning the full mass to the most frequent class, $y^{(n)}_k = 1$ for $k = \arg\max_j p^{(n)}_j$ and $y^{(n)}_k = 0$ otherwise, and minimizes the categorical cross-entropy
\begin{equation}
\label{eq:cce}
\mathcal{L}_{\mathrm{CCE}}^{(n)} = - \sum_{k=1}^{K} w_k \, y^{(n)}_k \log q^{(n)}_k ,
\end{equation}
with per-class weights $w_k$ set inversely proportional to the class frequency in the training set to counter class imbalance.

The ambiguity approach retains the full label distribution and trains the network to reproduce it. It minimizes the Kullback--Leibler divergence from the AGLD to the network~output,
\begin{equation}
\label{eq:kld}
\mathcal{L}_{\mathrm{KLD}}^{(n)} = D_{\mathrm{KL}}\!\left(\mathbf{p}^{(n)} \,\|\, \mathbf{q}^{(n)}\right) = \sum_{k=1}^{K} p^{(n)}_k \log \frac{p^{(n)}_k}{q^{(n)}_k} .
\end{equation}

Both objectives are optimized with Adam at an initial learning rate of $10^{-4}$ and a weight decay of $10^{-3}$ using mini-batches of 32 repetitions for at most 2500 epochs. The learning rate is halved after 10 epochs without an improvement in validation loss (minimum change $10^{-3}$), and training stops after 30 such epochs (minimum change $10^{-3}$). The checkpoint with the lowest validation loss is retained for evaluation. The implementation uses PyTorch (version 2.9.1). These hyperparameters were chosen empirically in our augmentation work~\citep{spilzBoostingAutomaticExercise2025} rather than formally optimized, and we keep them unchanged across all four datasets and both training objectives so that the results remain comparable to this previous work.

We evaluate both objectives in a five-fold cross-validation. The folds are stratified jointly by participant, class, and entropy category so that each fold holds a comparable share of every participant--class combination and, within each combination, a comparable proportion of unambiguous and ambiguous repetitions. {For this splitting operation we define a repetition as unambiguous if the entropy of its AGLD (Section~\ref{sec:entropy}) lies below an exemplary threshold of $0.47$~bit and ambiguous otherwise. This value corresponds to a distribution that assigns $0.9$ to one class and $0.1$ to another, which we take as the boundary of the unambiguous case. Here the threshold serves only to stratify the folds and the resulting fold assignment is shared by both training configurations, so it cannot bias their comparison.} The training and validation folds are augmented as described in~\citep{spilzBoostingAutomaticExercise2025}, with a 2:1 ratio of augmented to real examples. The test folds remain unaugmented.

\subsection{Evaluation Metrics}
\label{sec:metrics}

We compare the baseline and the ambiguity approach from Section~\ref{sec:training} along three~aspects: their classification performance, their ability to detect ambiguous repetitions, and their ability to identify the relevant classes of an ambiguous repetition. For a repetition $n$, $\mathbf{p}^{(n)}$ denotes its AGLD and $\mathbf{q}^{(n)}$ the softmax output of the trained network. {All three metrics take the AGLD as the reference and therefore measure how well an approach reproduces it. Whether the AGLD itself represents perceived ambiguity is a separate question, which we examine in Sections~\ref{sec:aglv-original} and \ref{sec:discussion-aglv}.}

Classification performance measures how well an approach recovers the single most relevant class of a repetition. We take the highest-probability class of the AGLD, $\arg\max_k p^{(n)}_k$, as the reference and the highest-probability class of the softmax output, $\arg\max_k q^{(n)}_k$, as the prediction and report the macro-averaged $F_1$-score across the $K$ classes of the respective dataset, denoted $F_1^{\mathrm{cls}}$, and the class-wise $F_{1,k}^{\mathrm{cls}}$.

Ambiguity detection measures how well an approach recovers from its softmax output whether a repetition is ambiguous. Following Section~\ref{sec:entropy}, a repetition is ambiguous when the entropy of its distribution exceeds a threshold $\tau$. We apply this criterion to both distributions so that the reference is positive when $H(\mathbf{p}^{(n)}) > \tau$ and the prediction is positive when $H(\mathbf{q}^{(n)}) > \tau$ and report the $F_1$-score of the predicted ambiguity against the reference ambiguity, denoted $F_1^{\mathrm{amb}}$.

The third aspect combines ambiguity detection with the identification of the relevant classes. For an ambiguous repetition the two highest-probability classes define which classes compete in its evaluation, with their ranking distinguishing the more likely class from the second. Let $\mathrm{top}_2(\cdot)$ denote the ordered pair of the two highest-probability classes of a distribution, sorted by decreasing probability. A reference repetition is positive when $H(\mathbf{p}^{(n)}) > \tau$, as above, and a prediction counts as a true positive only when the repetition is detected as ambiguous and $\mathrm{top}_2(\mathbf{q}^{(n)})$ equals $\mathrm{top}_2(\mathbf{p}^{(n)})$, which requires both classes to match in the same rank order. We report the $F_1$-score of this combined decision against the reference, denoted $F_1^{\mathrm{top2}}$.

Both entropy-based metrics, $F_1^{\mathrm{amb}}$ and $F_1^{\mathrm{top2}}$, depend on the threshold $\tau$, for which no established standard exists. We therefore evaluate them across the range of attainable entropy values, $\tau \in (0, \log_2 K)$, instead of fixing a single threshold.

The combined ambiguity and top-2 decision is reached over several stages, at each of which a repetition can be handled correctly or incorrectly. The aggregate scores $F_1^{\mathrm{amb}}$ and $F_1^{\mathrm{top2}}$ condense this into a single value and therefore do not reveal where the correct and incorrect decisions fall. To make them legible across the combined process, we additionally provide a hierarchical breakdown of the ambiguity detection and the top-2 identification at the {exemplary} threshold of $0.47$~bit introduced in Section~\ref{sec:training}. The breakdown follows the repetitions through four successive stages: whether a repetition is ambiguous, whether it is detected as ambiguous, whether its most relevant class is recovered, and whether its second relevant class is recovered.

\subsection{AGLD Versus Original Labels}
\label{sec:aglv-original}

The AGLD is the reference against which the trained network is evaluated in \mbox{Section~\ref{sec:metrics}}. This subsection relates the AGLD to the original labels of the datasets \mbox{(Section~\ref{sec:datasets})} along two aspects, its most frequent class and its entropy.

The first aspect is the most frequent class of the AGLD, $\arg\max_k p^{(n)}_k$. We compare it against the original label of each repetition and report the macro-averaged $F_1$-score across the $K$ classes of the respective dataset, denoted $F_1^{\mathrm{orig}}$, and the class-wise $F_{1,k}^{\mathrm{orig}}$. For RD and RGS the reference is the instruction-based label. For DS and HS, scored by three to five~raters, it is the majority vote across raters.

The second aspect is the entropy of the AGLD and applies to DS and HS, the datasets with several raters per repetition. A repetition is taken as ambiguous when the raters do not assign the same score and as unambiguous when they agree. We treat the AGLD entropy as the predictor and call a repetition ambiguous when $H(\mathbf{p}^{(n)}) > \tau$ (Section~\ref{sec:entropy}). RD and RGS hold a single instruction-based label per repetition (Section~\ref{sec:datasets}) and provide no comparable signal, so this comparison is limited to DS and HS. Across the range of attainable entropy values, here $\tau \in (0, \log_2 K)$ with $K = 3$ for the three classes of DS and HS, we report precision and recall of the predicted ambiguity against the rater-based reference.

\subsection{{Sensitivity Analysis of the Simulated Rater Variability $\sigma_c$}}
\label{sec:sensitivity}

{The standard deviations $\sigma_c$ represent the simulated rater variability as they control how strongly the labeling instances disagree by setting the spread of the sampled thresholds and thereby the entropy of the AGLD (Section~\ref{sec:aglv}). Since we set these values heuristically, we examine how sensitive the results are to this choice. For this purpose we generate two additional variants of the AGLD, one with all $\sigma_c$ halved and one with all $\sigma_c$ doubled, while the mean thresholds $\mu_c$, the criteria, and the number of labeling instances remain unchanged. The sensitivity analysis covers two parts that are carried out on different~datasets.}

{The first part repeats the comparison against the rater-based reference (Section~\ref{sec:aglv-original}). For both variants we report the precision and recall of the entropy-based ambiguity detection against the rater disagreement. This part is carried out on DS and HS, the only datasets scored by several raters, in order to draw on all the rater information that is available in our data.}

{The second part repeats the comparison of the baseline and the ambiguity approach (Section~\ref{sec:metrics}). For each variant we train both approaches on the AGLD of that variant, with the experimental design of Section~\ref{sec:training} unchanged, and evaluate $F_1^{\mathrm{cls}}$, $F_1^{\mathrm{amb}}$, and $F_1^{\mathrm{top2}}$. We restrict this part to RGS and DS, one dataset per exercise type, covering a foot drop exercise with a nominal class structure and an FMS task with an ordinal one because retraining both approaches for every variant multiplies the computational cost.}

\section{Results}
\label{sec:results}

This section reports the comparison of the baseline and the ambiguity approach along the three aspects introduced in Section~\ref{sec:metrics}: their classification performance, their detection of ambiguous repetitions, and their identification of the relevant classes of an ambiguous repetition across the four datasets. It then relates the AGLD to the original labels of the datasets. {Finally, it reports the sensitivity of these results to the simulated rater variability~$\sigma_c$.}

\subsection{Classification Performance}
\label{sec:results-classification}

Table~\ref{tab:classification-tratr} reports $F_1^{\mathrm{cls}}$ and the class-wise $F_1$-scores of the baseline and the ambiguity approach across the five folds, following the metric defined in Section~\ref{sec:metrics}. The ambiguity approach matches or exceeds the baseline $F_1^{\mathrm{cls}}$ on all four datasets.

On RD the two approaches reach the same $F_1^{\mathrm{cls}}$ ($0.92 \pm 0.02$), and the class-wise scores for CE, TL, SUP, and PRO differ by at most $0.01$. On RGS the $F_1^{\mathrm{cls}}$ is $0.89 \pm 0.02$ for the baseline and $0.90 \pm 0.02$ for the ambiguity approach, with the class-wise scores within $0.02$ of each other and the lowest value on IKF for both approaches ($0.82$ and $0.84$).

On DS the $F_1^{\mathrm{cls}}$ is higher for the ambiguity approach ($0.93$ to $0.97$). The difference is concentrated in FMS~2 ($0.86$ to $0.92$) and FMS~1 ($0.95$ to $0.98$), while FMS~3 reaches $1.00$ for both approaches. On HS the $F_1^{\mathrm{cls}}$ shows the largest difference ($0.69$ to $0.81$). It is concentrated in FMS~1 ($0.32$ to $0.63$), which also carries the widest fold-to-fold spread of all the class-wise scores ($\pm 0.18$ for the baseline and $\pm 0.13$ for the ambiguity approach). HS has the lowest $F_1^{\mathrm{cls}}$ and the highest spread among the four datasets.

\begin{table}[H]

\caption{Classification performance in the five-fold cross-validation. $F_1^{\mathrm{cls}}$ and class-wise $F_1$-scores (\mbox{mean $\pm$ standard} deviation over the five folds) for the baseline and the ambiguity approach on the four datasets. The class-wise columns $F_{1,1}^{\mathrm{cls}}$ to $F_{1,4}^{\mathrm{cls}}$ follow the class order per exercise, RD (CE, TL, SUP, and PRO), RGS (CE, IKF, IHA, and SH), and DS and HS (FMS~1, 2, and 3). DS and HS have only three classes, so $F_{1,4}^{\mathrm{cls}}$ is empty.\label{tab:classification-tratr}}

	\isPreprints{}{\begin{adjustwidth}{-\extralength}{0cm}}

		\begin{tabularx}{\widetablelength}{LlCCCCC}
\toprule
\textbf{Exercise} & \textbf{Setup} & \boldmath{$F_1^{\mathrm{cls}}$} & \boldmath{$F_{1,1}^{\mathrm{cls}}$} & \boldmath{$F_{1,2}^{\mathrm{cls}}$} & \boldmath{$F_{1,3}^{\mathrm{cls}}$} & \boldmath{$F_{1,4}^{\mathrm{cls}}$} \\
\midrule
\multirow[c]{2}{*}{RD} & Baseline & 0.92 $\pm$ 0.02 & 0.91 $\pm$ 0.02 & 0.90 $\pm$ 0.01 & 0.93 $\pm$ {0.02} & 0.94 $\pm$ 0.03 \\
 & Ambiguity Approach & 0.92 $\pm$ 0.02 & 0.92 $\pm$ 0.02 & 0.91 $\pm$ 0.02 & 0.93 $\pm$ 0.02 & 0.93 $\pm$ 0.03 \\
               \midrule
\multirow[c]{2}{*}{RGS} & Baseline & 0.89 $\pm$ {0.02} & 0.89 $\pm$ {0.06} & 0.82 $\pm$ 0.02 & 0.95 $\pm$ 0.03 & 0.89 $\pm$ 0.01 \\
 & Ambiguity Approach & 0.90 $\pm$ 0.02 & 0.91 $\pm$ {0.05} & 0.84 $\pm$ 0.02 & 0.96 $\pm$ 0.02 & 0.89 $\pm$ 0.01 \\
               \midrule
\multirow[c]{2}{*}{DS} & Baseline & 0.93 $\pm$ 0.01 & 0.95 $\pm$ 0.01 & 0.86 $\pm$ {0.03} & 1.00 $\pm$ 0.00 &  \\
 & Ambiguity Approach & 0.97 $\pm$ {0.02} & 0.98 $\pm$ 0.01 & 0.92 $\pm$ {0.04} & 1.00 $\pm$ 0.00 &  \\
               \midrule
\multirow[c]{2}{*}{HS} & Baseline & 0.69 $\pm$ 0.08 & 0.32 $\pm$ {0.18} & 0.91 $\pm$ {0.03} & 0.85 $\pm$ 0.05 &  \\
 & Ambiguity Approach & 0.81 $\pm$ {0.06} & 0.63 $\pm$ {0.13} & 0.93 $\pm$ 0.02 & 0.88 $\pm$ {0.05} &  \\

\bottomrule
\end{tabularx}
\isPreprints{}{\end{adjustwidth}}
\end{table}

\subsection{Ambiguity Detection Performance}
\label{sec:results-ambdet}

\begin{figure}[H]

\includegraphics[width=\textwidth]{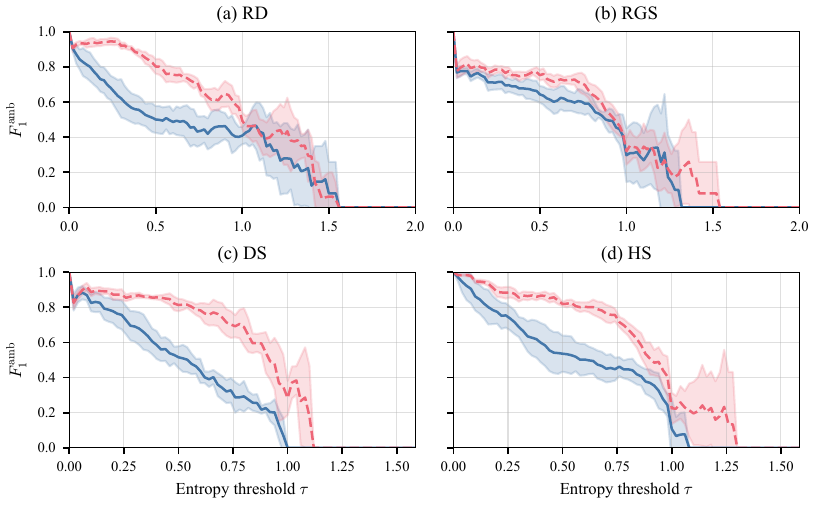}
\caption{Ambiguity detection performance $F_1^{\mathrm{amb}}$ as a function of the entropy threshold $\tau$ in the five-fold cross-validation, with one panel per dataset. Solid lines show the baseline; dashed lines show the ambiguity approach. Lines give the mean over the five folds and shaded bands the standard deviation. The threshold extends to the maximum attainable entropy of each dataset, $\log_2 4 = 2$~bit for RD and RGS and $\log_2 3 \approx 1.58$~bit for DS and HS.\label{fig:ambdet-threshold}}
\end{figure}

Figure~\ref{fig:ambdet-threshold} shows $F_1^{\mathrm{amb}}$ as a function of the entropy threshold $\tau$ for the baseline and the ambiguity approach on the four datasets, following the metric defined in Section~\ref{sec:metrics}.

\vspace{-4pt}

At both ends of the threshold axis the baseline and the ambiguity approach follow the same course across all four datasets. Near $\tau = 0$ both reach a near-perfect $F_1^{\mathrm{amb}}$ of about $1$. Above about $\tau = 1.0$ the $F_1^{\mathrm{amb}}$ of both approaches drops sharply and grows increasingly noisy, and the standard-deviation bands widen markedly on all four datasets.

In the intermediate range, from about $0.05$ to $1.0$~bit, the ambiguity approach reaches a clearly higher $F_1^{\mathrm{amb}}$ than the baseline on all four datasets. Within this range the $F_1^{\mathrm{amb}}$ of both approaches slowly declines as $\tau$ increases.

On RD and RGS the two approaches separate differently. For the baseline the RGS curve runs clearly above the RD curve across the intermediate range, where RD falls from low thresholds onward and reaches about $0.5$ by $\tau = 0.5$. For the ambiguity approach the order is reversed at first as RD stays near $0.9$ up to about $\tau = 0.3$ and leads RGS until about $\tau = 0.5$, beyond which the RD and RGS curves run close together.

On DS and HS the two datasets resemble each other within this range. The ambiguity curves of DS and HS follow a similar course, holding near $0.8$ on DS up to about $\tau = 0.6$ and near $0.85$ on HS into the mid-range. The baseline curves of the two datasets also run close together, declining to about $0.4$ over the same range.

\subsection{Ambiguity and Top-2 Detection Performance}
\label{sec:results-top2}

\begin{figure}[H]

\includegraphics[width=\textwidth]{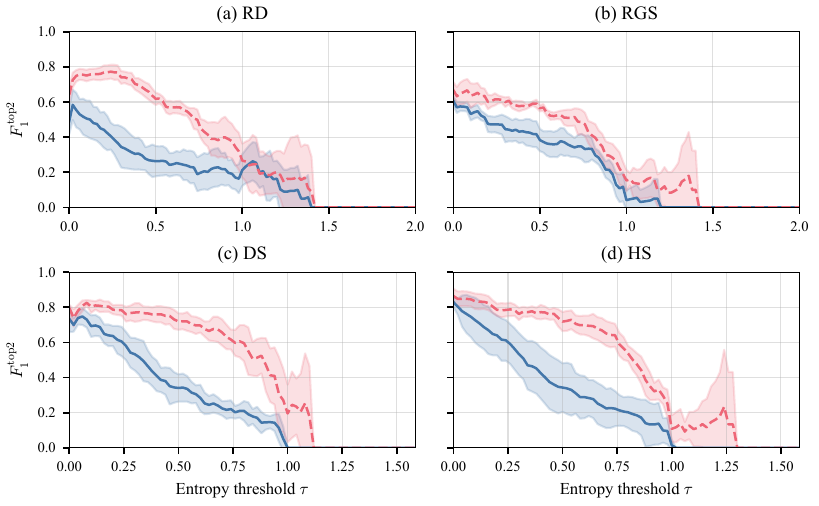}
\caption{Combined ambiguity detection and top-2 performance $F_1^{\mathrm{top2}}$ as a function of the entropy threshold $\tau$ in the five-fold cross-validation, with one panel per dataset. Solid lines show the baseline; dashed lines show the ambiguity approach. Lines give the mean over the five folds and shaded bands the standard deviation. The threshold extends to the maximum attainable entropy of each dataset, $\log_2 4 = 2$~bit for RD and RGS and $\log_2 3 \approx 1.58$~bit for DS and HS.\label{fig:top2-threshold}}
\end{figure}

Figure~\ref{fig:top2-threshold} shows $F_1^{\mathrm{top2}}$ as a function of the entropy threshold $\tau$ for the baseline and the ambiguity approach on the four datasets, following the metric defined in Section~\ref{sec:metrics}.

Most of the observations from $F_1^{\mathrm{amb}}$ (Figure~\ref{fig:ambdet-threshold}) carry over to $F_1^{\mathrm{top2}}$. In the intermediate range, from about $0.05$ to $1.0$~bit, the ambiguity approach reaches a clearly higher $F_1^{\mathrm{top2}}$ than the baseline on all four datasets, and, within this range, the $F_1^{\mathrm{top2}}$ of both approaches declines slowly as $\tau$ increases. On DS and HS the two datasets resemble each other within this range, with the baseline curves of the two following a similar course and the ambiguity curves doing the same. On RD and RGS the baseline runs higher on RGS than on RD, while, for the ambiguity approach, RD leads at first and the RD and RGS curves run close together beyond about $\tau = 0.5$. Above about $\tau = 1.0$ the same behavior over all the datasets returns as the curves grow strongly noisy and the standard-deviation bands widen.

The behavior near $\tau = 0$ differs from $F_1^{\mathrm{amb}}$. Where $F_1^{\mathrm{amb}}$ starts near $1$, $F_1^{\mathrm{top2}}$ starts clearly lower on all four datasets, at about $0.7$ to $0.8$. Across the threshold range the $F_1^{\mathrm{top2}}$ curves also lie below their $F_1^{\mathrm{amb}}$ counterparts. The baseline curves run about $0.2$ lower and the ambiguity curves about $0.15$ lower on RD and RGS and about $0.1$ lower on DS and HS.
\vspace{-4pt}

\subsection{Hierarchical Outcome Breakdown}
\label{sec:results-tree}

\begin{figure}[H]

\hspace{-18pt}\includegraphics[width=\textwidth]{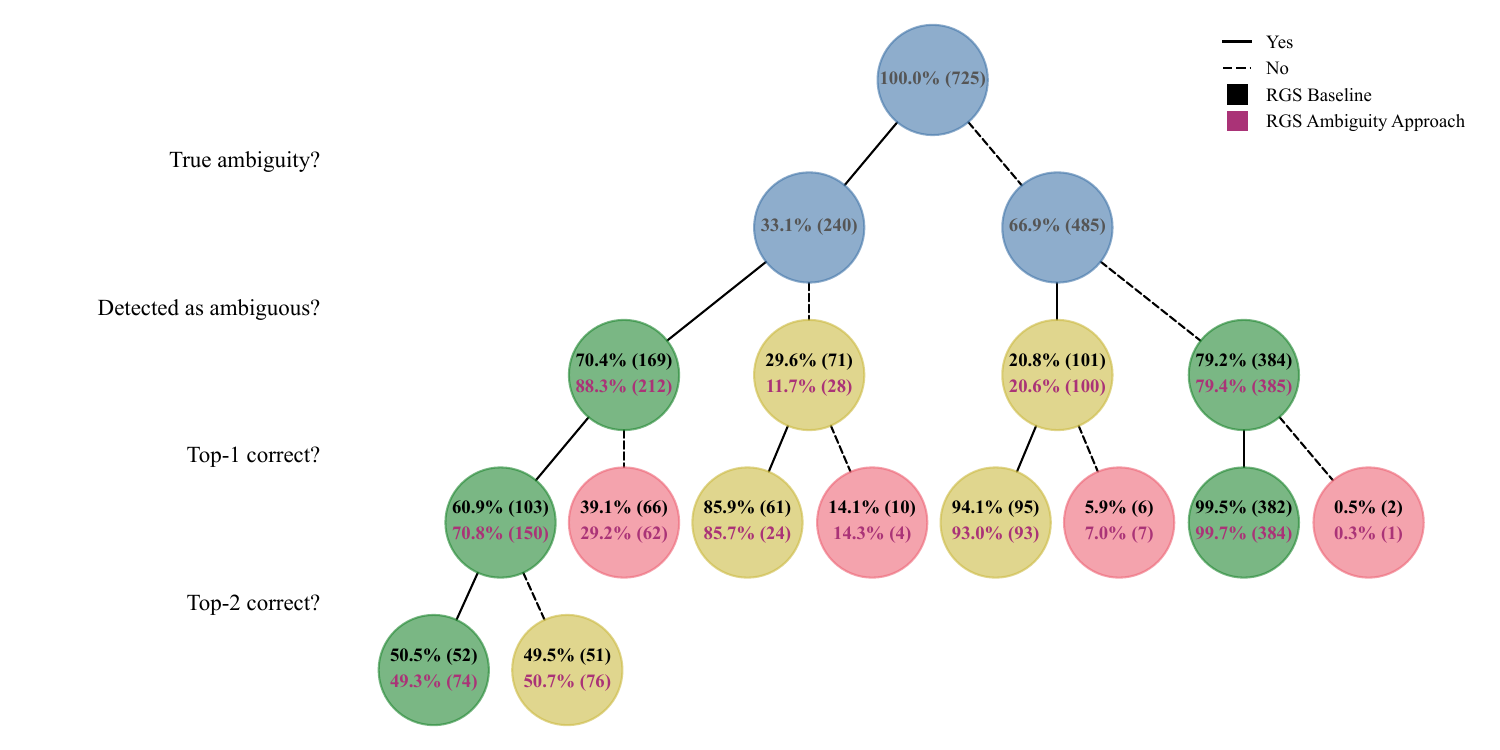}
\caption{Outcome tree for RGS at the {exemplary} entropy threshold $\tau = 0.47$~bit (Section~\ref{sec:training}). Solid lines mark the Yes branch and dashed lines the No branch. Each node gives the share of its parent with the absolute count in parentheses. The root and the two nodes of the ambiguity stage are shown in blue. They follow from the AGLD alone and therefore carry a single value that applies to both approaches. From the detection stage onward each node reports the baseline (black) and the ambiguity approach (magenta). The terminal leaves are colored by outcome category: green for a fully correct decision, red for an incorrect one, and yellow for an intermediate case in which the ambiguity decision is wrong, a false positive or a false negative, while the most relevant class is still recovered. The final decision is shown for the leftmost path, where a repetition is ambiguous, detected as such, top-1 and top-2 correct.\label{fig:tree-rgs}}
\end{figure}

Figure~\ref{fig:tree-rgs} shows the outcome tree for RGS at the {exemplary} entropy threshold \mbox{$\tau = 0.47$~bit}, following the breakdown defined in Section~\ref{sec:metrics}. We show RGS as a four-class dataset. The trees for RD, DS, and HS are given in Appendix~\ref{app:outcome-trees}.

Each leaf of the tree is colored by the type of outcome it represents. Green marks a fully correct decision and red an incorrect one. Between the two, yellow marks an intermediate outcome in which the ambiguity decision is wrong, a false positive or a false negative, while the most relevant class is still recovered correctly.

We read each tree along the four branches that combine the reference and the detected ambiguity at the detection stage: true positives and false negatives among the ambiguous repetitions, false positives and true negatives among the unambiguous ones.

On the true-positive branch the ambiguity approach detects a larger share of the ambiguous repetitions than the baseline on all four datasets. Among these detected repetitions the most relevant class is recovered more often by the ambiguity approach than by the baseline, while the second relevant class is recovered to a similar degree under both approaches. This second-class recovery is higher on DS and HS than on RD and RGS independent of the approach. On the false-negative branch, the ambiguous repetitions not detected as such, fewer repetitions go undetected under the ambiguity approach than under the baseline, and the most relevant class is recovered well by both approaches.

On the false-positive branch the difference between the approaches varies by dataset. The ambiguity approach flags a larger share of the unambiguous repetitions than the baseline on RD and HS and a comparable share on RGS and DS. Across these false positives the recovery of the most relevant class is comparable between the two approaches or favors the ambiguity approach depending on the dataset. The true-negative branch is the complement of the false positives, so the two approaches run comparably on RGS and DS while the baseline retains the larger share on RD and HS. The recovery of the most relevant class on this branch is comparable across all four datasets.
\vspace{-6pt}

\subsection{AGLD Versus Original Labels}
\label{sec:results-aglv-original}

We report the two comparisons defined in Section~\ref{sec:aglv-original}: the first relates the highest-probability class of the AGLD to the original label of a repetition and the second relates the AGLD entropy to the rater disagreement on DS and HS.

The results of the first comparison are reported in Table~\ref{tab:aglv-original} as $F_1^{\mathrm{orig}}$ and its class-wise scores. The $F_1^{\mathrm{orig}}$ is highest on RD at $0.91$ and more moderate on the other three datasets, at $0.83$ on RGS, $0.84$ on DS, and $0.76$ on HS, the lowest of the four.

\begin{table}[H]
\caption{Agreement between the highest-probability class of the AGLD and the original labels. $F_1^{\mathrm{orig}}$ and class-wise $F_{1,k}^{\mathrm{orig}}$ on the four datasets, computed over all the repetitions. The class-wise columns $F_{1,1}^{\mathrm{orig}}$ to $F_{1,4}^{\mathrm{orig}}$ follow the class order listed for each exercise. DS and HS have three classes, so $F_{1,4}^{\mathrm{orig}}$ is empty.\label{tab:aglv-original}}
\begin{tabularx}{\textwidth}{cCCCCC}
\toprule
\textbf{Exercise} & \boldmath{$F_1^{\mathrm{orig}}$} & \boldmath{$F_{1,1}^{\mathrm{orig}}$} & \boldmath{$F_{1,2}^{\mathrm{orig}}$} & \boldmath{$F_{1,3}^{\mathrm{orig}}$} & \boldmath{$F_{1,4}^{\mathrm{orig}}$} \\

\midrule
\shortstack{RD\\{\small(CE / TL / SUP / PRO)}} & 0.91 & 0.90 & 0.87 & 0.92 & 0.93 \\\midrule
\shortstack{RGS\\{\small(CE / IKF / IHA / SH)}} & 0.83 & 0.79 & 0.73 & 0.95 & 0.84 \\\midrule
\shortstack{DS\\{\small(FMS~1 / 2 / 3)}} & 0.84 & 0.89 & 0.64 & 1.00 &  \\\midrule
\shortstack{HS\\{\small(FMS~1 / 2 / 3)}} & 0.76 & 0.63 & 0.87 & 0.78 &  \\
\bottomrule
\end{tabularx}
\end{table}

On DS and HS the class-wise scores reveal a pattern once they are read together with the rater disagreement per class (Figure~\ref{fig:class-dist}). For these two datasets, where several raters scored each repetition, the figure gives per class the count of repetitions on which the raters agreed and the count on which they disagreed (agree/disagree). We take the disagreement share of a class as its disagree count relative to its class total. Ordering the classes by this share reverses their order by class-wise $F_1^{\mathrm{orig}}$ (Table~\ref{tab:aglv-original}). On DS the disagreement share rises from $0\%$ on FMS~3 through $9\%$ on FMS~1 to $30\%$ on FMS~2, while $F_1^{\mathrm{orig}}$ falls from $1.00$ through $0.89$ to $0.64$ in the same order. On HS it rises from $27\%$ on FMS~2 through $31\%$ on FMS~3 to $38\%$ on FMS~1, while $F_1^{\mathrm{orig}}$ falls from $0.87$ through $0.78$ to $0.63$.

The results of the second comparison are shown in Figure~\ref{fig:aglv-recall-precision} as the recall and precision of the AGLD-entropy-based ambiguity detection against the rater-based reference as a function of the entropy threshold $\tau$. {We describe here the main configuration $1\,\sigma_c$.} On both datasets the recall starts near $1.0$ and declines as $\tau$ increases, while the precision starts below $0.3$ and rises gradually to about $0.4$. The recall and precision follow a similar course on DS and HS across the low-to-mid range. Beyond about $\tau = 1.0$ the recall falls to near zero and the precision becomes noisy. {The figure additionally shows the two variants of the sensitivity analysis, $0.5\,\sigma_c$ and $2\,\sigma_c$, whose results are reported in Section~\ref{sec:results-sensitivity}.}

\subsection{{Sensitivity to the Simulated Rater Variability $\sigma_c$}}
\label{sec:results-sensitivity}

\begin{figure}[H]
\centering
\includegraphics[width=\textwidth]{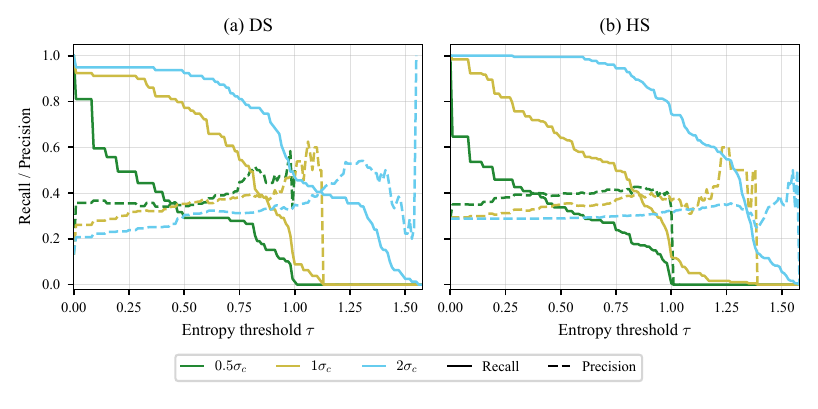}
\caption{{Recall and precision of the AGLD-entropy-based ambiguity detection against the rater-based reference as a function of the entropy threshold $\tau$. A repetition is positive in the reference when its raters do not assign the same score and positive in the prediction when $H(\mathbf{p}^{(n)}) > \tau$ (Section~\ref{sec:aglv-original}). Solid lines show recall; dashed lines show precision. Colors distinguish the main configuration $1\,\sigma_c$ from the two variants of the sensitivity analysis, $0.5\,\sigma_c$ and $2\,\sigma_c$ (Section~\ref{sec:sensitivity}). The threshold extends to the maximum attainable entropy of the three-class datasets, $\log_2 3 \approx 1.58$~bit.}\label{fig:aglv-recall-precision}}
\end{figure}

{This section reports the results of the sensitivity analysis defined in Section~\ref{sec:sensitivity}. The first part repeats the comparison of the AGLD entropy against the rater-based reference for variants $0.5\,\sigma_c$ and $2\,\sigma_c$, which Figure~\ref{fig:aglv-recall-precision} shows alongside the main configuration $1\,\sigma_c$.}

{Both variants follow the qualitative course described for $1\,\sigma_c$ in Section~\ref{sec:results-aglv-original}, with the recall declining as $\tau$ increases and the precision running at a similar level across the low-to-mid range on both datasets. The variants differ from the main configuration in how the recall declines along the threshold axis. Across the low-to-mid range the recall curves are ordered by the variant on both datasets, with $2\,\sigma_c$ running highest, $1\,\sigma_c$ in between, and $0.5\,\sigma_c$ lowest. On DS, for example, the recall of $0.5\,\sigma_c$ falls below $0.5$ by about $\tau = 0.25$, whereas $2\,\sigma_c$ stays above $0.9$ up to about $\tau = 0.5$, and HS shows the same ordering.}

{The second part repeats the comparison of the baseline and the ambiguity approach for the two variants on RGS and DS. The corresponding table and figures are provided in Appendix~\ref{app:sensitivity}, and we describe them in the order of the three metrics (Sections~\ref{sec:results-classification}--\ref{sec:results-top2}).}

{Table~\ref{tab:sensitivity-classification} reports the classification performance for all three variants. Across the variants $F_1^{\mathrm{cls}}$ varies by at most $0.02$ per approach and dataset, and the ambiguity approach matches or slightly exceeds the baseline in every variant on RGS ($0.88$ to $0.89$ for the baseline against $0.90$ to $0.91$ for the ambiguity approach) and DS ($0.93$ to $0.94$ against $0.95$ to $0.97$).}

{Figure~\ref{fig:sensitivity-ambdet} shows $F_1^{\mathrm{amb}}$ as a function of the entropy threshold $\tau$ for the three variants. The course described for $1\,\sigma_c$ in Section~\ref{sec:results-ambdet} holds for every variant, with a near-perfect $F_1^{\mathrm{amb}}$ near $\tau = 0$, increasingly noisy curves at high thresholds, and the ambiguity approach above the baseline across the intermediate range on both datasets. Within this range the three baseline curves run close together, with $1\,\sigma_c$ and $2\,\sigma_c$ nearly coinciding and $0.5\,\sigma_c$ slightly below. The ambiguity curves, in contrast, are ordered by the variant and rise with $\sigma_c$, with the $2\,\sigma_c$ curve holding high values throughout the range. The separation between the two approaches therefore widens as $\sigma_c$ increases.}

{Figure~\ref{fig:sensitivity-top2} shows the corresponding $F_1^{\mathrm{top2}}$ curves. As in Section~\ref{sec:results-top2}, they keep the shape of their $F_1^{\mathrm{amb}}$ counterparts and run below them, and this holds for every variant on both datasets, so the ordering of the curves carries over.}

\section{Discussion}
\subsection{Classification Performance}
\label{sec:discussion-classification}

Across the four datasets the ambiguity approach matches or exceeds the $F_1^{\mathrm{cls}}$ of the baseline (Table~\ref{tab:classification-tratr}). On RD, RGS, and DS the two approaches reach the same value or the ambiguity approach is slightly better, while HS shows a clear improvement ($0.69$ to $0.81$).

However, the gain on HS is concentrated in FMS~1 ($0.32$ to $0.63$), which accounts for the largest share of the macro-level improvement. This class contains only eight repetitions in the dataset (Figure~\ref{fig:class-dist}), the fewest of any class across all four datasets, and shows the widest fold-to-fold spread of all the class-wise scores ($\pm 0.18$ for the baseline and $\pm 0.13$ for the ambiguity approach). The observed improvement therefore rests on too few cases to support a reliable conclusion.

{Independently of this,} the comparison shows that representing ambiguity in the training target does not reduce the classification performance on the most relevant class.

\subsection{Ambiguity Detection and Top-2 Detection}
\label{sec:discussion-ambdet}

Across the low-to-mid threshold range, from about $0.05$ to $1.0$~bit, the ambiguity approach reaches a higher $F_1^{\mathrm{amb}}$ than the baseline on all four datasets (Figure~\ref{fig:ambdet-threshold}). Since the two approaches differ only in the training target, this separation follows from the distribution-valued target in combination with the KLD objective, which trains the network to reproduce the full AGLD and thereby to recover the ambiguity of a repetition from its~output.

This separation is only informative away from the two ends of the threshold axis. Near $\tau = 0$ almost every repetition clears the threshold and counts as ambiguous, in the reference as well as in the prediction. The $F_1^{\mathrm{amb}}$ is then near-perfect because almost everything is labeled positive, not because the approaches locate the ambiguous repetitions well. At the other end, above about $1.0$~bit, only a few repetitions carry such high entropy (Figure~\ref{fig:entropy-dist}), so the positives grow sparse and the curves turn noisy. The comparison therefore holds only in the intermediate range named above.

Both approaches share a further trend within this range: the $F_1^{\mathrm{amb}}$ declines as $\tau$ rises, which follows from how the threshold interacts with the predicted entropy. At a low $\tau$ the predicted entropy $H(\mathbf{q}^{(n)})$ can deviate substantially from the reference entropy $H(\mathbf{p}^{(n)})$ and still stay on the same side of the threshold, so the detection tolerates a large error. As $\tau$ rises towards the entropy of the repetitions, this tolerance shrinks, so a small deviation increasingly moves the prediction to the wrong side and the $F_1^{\mathrm{amb}}$ falls.

The same picture holds for the combined ambiguity and top-2 decision $F_1^{\mathrm{top2}}$ (Figure~\ref{fig:top2-threshold}). Across the low-to-mid range the ambiguity approach reaches a higher $F_1^{\mathrm{top2}}$ than the baseline on all four datasets, so, beyond detecting ambiguity, it also identifies the two~relevant classes of an ambiguous repetition more reliably. This is expected since the ambiguity approach learns the full AGLD and with it the relative weight of every class, and the top-2 decision reads off exactly the two highest-weighted classes.

Across the threshold range the $F_1^{\mathrm{top2}}$ keeps the shape of the corresponding $F_1^{\mathrm{amb}}$ but runs below it (Figures~\ref{fig:ambdet-threshold} and \ref{fig:top2-threshold}).
This offset follows from the construction of the metric, which extends the ambiguity detection by a second condition. A repetition now counts as correct only when it is detected as ambiguous and its two highest-probability classes match those of the AGLD. This second condition can only reject repetitions that the detection alone would have accepted, never add new ones, so the curve stays below $F_1^{\mathrm{amb}}$. The reduction relative to $F_1^{\mathrm{amb}}$ is larger for the ambiguity curves on RD and RGS, at about $0.15$, than on DS and HS, at about $0.1$. We hypothesize that this difference reflects the number of classes. The second condition is evaluated only after the most relevant class is already correct, so the second relevant class is selected among the $K - 1$ remaining classes, three on the four-class datasets, RD and RGS, and two on the three-class datasets, DS and HS. The wider set of candidate classes on RD and RGS leaves more room for the second class to be misidentified, which enlarges this gap.

The results so far hold across all four datasets, but the size of the advantage of the ambiguity approach over the baseline differs between them. For $F_1^{\mathrm{amb}}$ this advantage is widest on RD, DS, and HS and narrowest on RGS (Figure~\ref{fig:ambdet-threshold}). The narrow gap on RGS is not mainly a property of the ambiguity approach. Its curve runs below RD up to about $\tau = 0.5$ and matches RD across the remaining range to about $\tau = 1.0$, so it trails the other datasets only over the lower half of the range. We attribute the narrow gap instead to the RGS baseline, which reaches a clearly higher $F_1^{\mathrm{amb}}$ than the baseline on the other three datasets. The stronger baseline leaves less room for the ambiguity approach to improve on it, so the separation comes out smallest on RGS even though the ambiguity approach itself performs comparably to the other datasets.

The outcome trees allow a more differentiated view of the correct and incorrect decisions than the aggregate curves as they resolve the combined decision into its successive stages (Figure~\ref{fig:tree-rgs} and Appendix~\ref{app:outcome-trees}). Within them we marked a third category in yellow, the repetitions whose ambiguity is misjudged (false positive or false negative), while the most relevant class is still recovered. We regard these as less consequential errors as the reported class remains closer to the actual one. Feedback that reports FMS~1 instead of FMS~1 and 2 is less wrong than feedback that reports FMS~3 for a repetition that is actually FMS~1.

At the threshold $\tau = 0.47$~bit the trees show that the ambiguity approach detects ambiguous repetitions more sensitively than the baseline, in part at the expense of specificity on RD and HS. On these wrongly flagged repetitions, however, the recovery of the most relevant class remains good, so, by the criterion introduced above, they fall into the less consequential category of errors.

A further observation concerns the true-positive branch, where the most relevant class is recovered more often by the ambiguity approach than by the baseline. On these repetitions the approach therefore does not only detect the ambiguity but also infers the leading class more reliably. We hypothesize that this follows from the training signal on the ambiguous repetitions. Under one-hot labels two near-identical borderline repetitions are mapped to conflicting single classes, whereas the distribution-valued target assigns them comparable distributions and thereby resolves this contradiction, which makes the leading class easier to recover.

\textls[15]{Across both detection metrics the comparison is consistent. The ambiguity approach improves the detection of ambiguous repetitions and the identification of their relevant classes, while the classification of the most relevant class stays at least on par with the baseline (Section~\ref{sec:discussion-classification}). Representing the label distribution in the training target therefore yields more information about the ambiguity of a repetition at no cost to classification~performance.}

\subsection{Reasonable Entropy Thresholds}
\label{sec:discussion-threshold}

In this work we evaluated the ambiguity detection across the whole range of attainable entropy thresholds so that the reader sees the full behavior rather than a single chosen operating point. A practical deployment, however, has to commit to one threshold $\tau$ at which a repetition is called ambiguous. To the best of our knowledge no established value exists for this choice, so we share the considerations that we find to be relevant for it.

For the outcome trees we already fixed $\tau$ heuristically at {the exemplary value} $0.47$~bit (Section~\ref{sec:training}). The $F_1^{\mathrm{amb}}$ and $F_1^{\mathrm{top2}}$ curves place this value in the range where the detection performs among its best on all the datasets. The optimum itself, however, falls at a different $\tau$ for each exercise, as the individual graphs show. Reading a threshold off these optima is not straightforward. The distribution of repetitions over the entropy is strongly dataset-specific (Figure~\ref{fig:entropy-dist}), and the distance between a repetition's entropy and $\tau$ sets how far the network may misestimate before the decision flips, which favors lower thresholds (Section~\ref{sec:discussion-ambdet}). The per-exercise optimum therefore does not translate into a generally sensible threshold. 

\begin{figure}[H]
\centering
\includegraphics[width=0.66\textwidth]{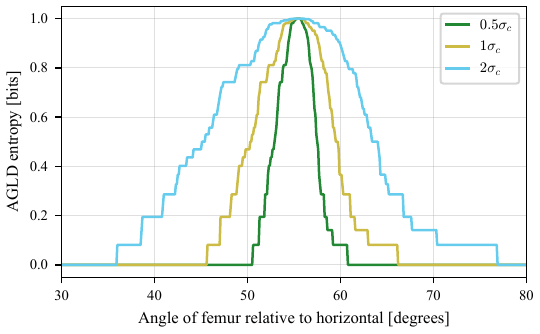}
\caption{Relation between the value a repetition reaches on one example criterion of the DS exercise and the resulting AGLD entropy{, with colors distinguishing the main configuration $1\,\sigma_c$ from the two~variants of the sensitivity analysis, $0.5\,\sigma_c$ and $2\,\sigma_c$ (Sections~\ref{sec:aglv} and \ref{sec:sensitivity})}. The criterion is the angle of the femur relative to the horizontal at the lowest point of the squat. Entropy is given in bits. Each labeling instance treats the criterion as fulfilled or violated by comparing this angle against its own threshold, drawn from $\mathcal{N}(\mu_c, \sigma_c^2)$, with the mean near $56^\circ$. {For $1\,\sigma_c$ the} entropy stays near zero while the angle lies well away from the mean and rises to its maximum as the angle approaches the mean from either side, with the transition spanning about $10^\circ$ on each side. {A larger $\sigma_c$ widens this transition zone and a smaller $\sigma_c$ narrows it.} The maximum stays near $1$~bit rather than the three-class maximum of $\log_2 3 \approx 1.58$~bit as only two classes compete at this single criterion.\label{fig:entropy-criterion}}

\end{figure}

A further observation helps to bound this choice. Figure~\ref{fig:entropy-criterion} relates the entropy of a repetition to the value it reaches on one example criterion of the DS exercise, the angle of the femur relative to the horizontal. The shape of this relation reflects the mean and standard deviation we set for the rater threshold distributions (Section~\ref{sec:aglv}). {Under the main configuration $1\,\sigma_c$, while} the angle lies well away from the mean threshold near $56^\circ$, the raters agree and the entropy stays near zero. As the angle approaches the threshold from either side, the entropy rises to its maximum, and the whole transition spans only about $10^\circ$ on each side. The rise is steep. Between $48^\circ$ and $52^\circ$ the entropy already doubles across a difference of just $4^\circ$. Such a difference is barely visible to an observer, who would likely read both $48^\circ$ and $52^\circ$ as close enough to the threshold to count as ambiguous. We conclude from this that the exact value of $\tau$ matters less than the choice to keep it low. A low $\tau$ is the safer setting since even a small disagreement among the raters already places a repetition near a criterion threshold and therefore at the boundary between two classes. {Where a deployed system commits to a single operating point, this choice amounts to a calibration per exercise. $\tau$ has to be set relative to the entropy distribution the repetitions of that exercise carry, with the considerations above favoring the lower end of the range. The two variants of the simulated rater variability that Figure~\ref{fig:entropy-criterion} shows alongside $1\,\sigma_c$ shift this entropy distribution, and Section~\ref{sec:discussion-sensitivity} discusses what follows from that for the choice of $\tau$.}

\subsection{Validity of the AGLD}
\label{sec:discussion-aglv}

The AGLD serves as the reference for every experiment in this work, so its validity as a representation of the data determines how much the results reported above can carry. Having compared the two training objectives against it, we now examine the reference itself. We consider two aspects: whether the most probable class of an AGLD agrees with the original single label of a repetition and whether the AGLD entropy reflects the ambiguity that human raters perceive.

The first aspect is the $F_1^{\mathrm{orig}}$ comparison in Section~\ref{sec:results-aglv-original}, which checks whether the highest-probability class of the AGLD coincides with the original single label of a repetition. The agreement is high on RD at $0.91$ and more moderate on the remaining datasets, at $0.83$ on RGS, $0.84$ on DS, and $0.76$ on HS (Table~\ref{tab:aglv-original}). These values are reasonably high and indicate that the highest-probability class of the AGLD recovers the original label well.

This agreement is partly built in. The labeling thresholds were fitted in our augmentation work to maximize the agreement between the one-hot automatic labels and the original labels (Section~\ref{sec:aglv}), and the AGLDs use the same threshold set as the mean of their sampling distributions. The informative question is therefore not whether the labels agree but whether extending the labeling to a full distribution preserves this agreement. It does as the discrepancy with the original labels was already present in the one-hot labeling, which reached agreement at a per-exercise level that is comparable to the AGLD mode here~\citep{spilzBoostingAutomaticExercise2025}. For the validity of the AGLD, the first aspect is thus a consistency check rather than an independent validation.

The agreement is nonetheless not perfect. We hypothesize that the misclassified repetitions are predominantly ambiguous ones{ since a small change in the movement or in a labeling threshold can swap the two~highest-probability classes of such a repetition.} Measured against a single original label, {this} swap counts as a full error, although the two~classes were nearly tied. This link is visible at the class level for DS and HS, where rater disagreement provides a per-class measure of ambiguity. The order of the classes by their disagreement share is the inverse of their order by class-wise $F_{1,k}^{\mathrm{orig}}$ (Section~\ref{sec:results-aglv-original}). The classes on which the raters disagree more are thus the classes the highest-probability class of the AGLD recovers less well, as expected if the deviations originate in ambiguous repetitions. 

The second aspect concerns whether the AGLD entropy tracks the ambiguity that human raters perceive. {Assessing this requires an external reference for the perceived ambiguity, which only DS and HS provide through the disagreement of their raters. What distinguishes the two types of datasets is therefore not the generation of the AGLD, which is identical for all four (Section~\ref{sec:aglv}), but the reference that is available to examine it.} The comparison below draws on the relation between the entropy and the rater disagreement in DS and HS (Figure~\ref{fig:aglv-recall-precision}).

A first observation is the low precision on both datasets. We attribute this to how the reference is derived: the raters were never asked whether a repetition is ambiguous, so we read ambiguity from their disagreement, and, while disagreement reliably marks a repetition, agreement does not rule it out as raters can agree on a score and still have perceived a borderline case. Many repetitions that the entropy marks as ambiguous therefore enter the reference as non-ambiguous, some of them possibly ambiguous after all. Because the raters' perception of ambiguity was never recorded, these cases cannot be separated from genuine false detections, so the precision understates how well the entropy locates ambiguous repetitions.

A second observation is that the recall declines as $\tau$ rises. We attribute this to the reference not capturing how ambiguous a repetition is, whereas the entropy resolves these degrees. We hypothesize that, at higher $\tau$, a growing number of repetitions carry a noticeable entropy that no longer clears the threshold. Their ambiguity is still visible in the entropy, but the threshold excludes them, and, since the reference marks every disagreement as ambiguous without distinguishing levels, these repetitions enter as false negatives. The count of false negatives therefore rises with $\tau$ by construction.

A last point concerns the behavior at both ends of the $\tau$ axis, which carries little meaning for the reasons given in Section~\ref{sec:discussion-ambdet}. The informative comparison therefore lies in the intermediate range.

Taken together, these evaluations support a limited conclusion. The AGLDs appear to lie close to the perceived ambiguity, and the two datasets follow the same pattern, suggesting that the situation is comparable across exercises. The reference, however, prevents an exact assessment for the reasons given above. A dataset in which the raters mark each repetition as ambiguous or unambiguous directly, rather than only assigning a score, would remove this limitation and allow a precise evaluation.

\subsection{{Sensitivity to the Simulated Rater Variability $\sigma_c$}}
\label{sec:discussion-sensitivity}

{The sensitivity analysis (Sections~\ref{sec:sensitivity} and \ref{sec:results-sensitivity}) shows that the central comparisons of this work do not depend on the specific value of $\sigma_c$. With the standard deviations halved or doubled, the classification performance remains within $0.02$ of the main configuration and the ambiguity approach matches or exceeds the baseline in every variant (Table~\ref{tab:sensitivity-classification}). The ambiguity approach also reaches higher $F_1^{\mathrm{amb}}$ and $F_1^{\mathrm{top2}}$ than the baseline across the intermediate threshold range in every variant (Figures~\ref{fig:sensitivity-ambdet} and \ref{fig:sensitivity-top2}).}

{What does depend on $\sigma_c$ is how much entropy the repetitions carry. A larger $\sigma_c$ widens the spread of the sampled thresholds, so the labeling instances disagree on more repetitions and the AGLD entropies rise across the dataset, while a smaller $\sigma_c$ narrows this spread and lowers them, as Figure~\ref{fig:entropy-criterion} illustrates for one example criterion (Section~\ref{sec:discussion-threshold}). At a fixed threshold $\tau$, a larger share of repetitions therefore count as ambiguous under $2\,\sigma_c$ and a smaller share under $0.5\,\sigma_c$. This explains the recall ordering in the comparison against the rater-based reference (Section~\ref{sec:results-sensitivity}). The reference is fixed by the rater disagreement, while, under $2\,\sigma_c$, more repetitions clear a given threshold, so the recall stays high over a wider range of $\tau$, and under $0.5\,\sigma_c$ it falls off earlier.}

{We hypothesize that the same shift contributes to the detection scores of the ambiguity approach through two mechanisms. First, at a fixed $\tau$, the higher entropies under a larger $\sigma_c$ lie farther above the threshold, so the predicted distribution can deviate more strongly from the AGLD before the decision flips, which is the tolerance effect described in Section~\ref{sec:discussion-ambdet}. Second, the $F_1$-score does not account for true negatives. As a larger $\sigma_c$ lifts the entropy of all the repetitions, the share of positives grows at any fixed $\tau$, and the $F_1^{\mathrm{amb}}$ rises with this base rate even without better discrimination. Both mechanisms reverse under $0.5\,\sigma_c$, where the entropies sit closer to the threshold and the share of positives shrinks, which is consistent with its curves running lowest. The absolute detection scores are therefore not directly comparable across the variants.}

{The practical consequence is that $\sigma_c$ and $\tau$ are coupled. Since $\sigma_c$ sets how much entropy the repetitions carry, a fixed $\tau$ corresponds to a different operating point under each variant. For deployment it is therefore not the exact value of $\sigma_c$ that requires calibration, provided that it is set to a plausible magnitude, but the threshold $\tau$ relative to the entropy distribution this $\sigma_c$ induces. Within the tested range the qualitative conclusions remain unchanged, and the considerations for choosing a low $\tau$ (Section~\ref{sec:discussion-threshold}) apply to every variant.}

\subsection{Limitations}
\label{sec:discussion-limitations}

Several limitations qualify these results. {The scope of the evaluation treats ambiguity as a two-class phenomenon. The AGLD represents the full distribution over all the classes, but the top-2 metric and the $0.9$ to $0.1$ split behind the exemplary threshold (Section~\ref{sec:training}) both target two competing classes, so the higher-order ambiguity of a repetition in which three or more classes compete is not captured. The evaluation could be extended to such repetitions through additional entropy thresholds that distinguish levels of ambiguity since the entropy rises with the number of competing classes, together with a top-$k$ extension of the top-2 read-out. These thresholds would have to be calibrated in the same way as $\tau$, so we regard this extension as future work.}

{The AGLD itself remains a simplified model of rater behavior. It reproduces only the threshold variation between raters established in Sections~\ref{sec:introduction} and \ref{sec:aglv}, so each labeling instance acts as an idealized rater, one who knows the evaluation criteria, observes the repetition from an ideal perspective with access to the exact kinematic quantities, applies its thresholds exactly and repeatably, and differs from the other instances only in these thresholds. Further properties of rater behavior, such as experience and observation perspective, are not represented. Relaxing the repeatability of the thresholds within an instance could include them, with experience entering through thresholds that vary across the repetitions that an instance rates, an experienced rater corresponding to a constant threshold, and the observation perspective entering through the spread of this variation, set higher for an unfavorable perspective. We regard such refinements as future work. How far these simplified distributions reflect perceived ambiguity can only be examined indirectly and only for DS and HS (Section~\ref{sec:discussion-aglv}), so, for RD and RGS, the ambiguity representation rests on the construction alone.}

{Beyond the model of rater behavior, the values of the standard deviations $\sigma_c$ and the normal form of the threshold distributions are uncalibrated assumptions, set per type of kinematic quantity (Section~\ref{sec:aglv}). The sensitivity analysis shows that the qualitative conclusions do not depend on the specific $\sigma_c$ within the tested range (Section~\ref{sec:discussion-sensitivity}). Calibrating the $\sigma_c$ and the operating point $\tau$ against perceived ambiguity, and testing the normality assumption, would, however, require the reference described in Section~\ref{sec:discussion-aglv}, which our data does not provide, so these evaluations remain for future work.}

{The AGLDs also reduce the effort of obtaining a label distribution without removing it. They replace a large rater pool with repeated runs of the rule-based labeling procedure, yet this procedure still has to be implemented for each exercise, which requires a qualified rater to translate the exercise into criteria and the sensor setup and inverse-kinematics simulation of Section~\ref{sec:aglv} to compute the underlying kinematic quantities. Compared to a multi-human-rater setup, this method should nevertheless be more efficient in most cases.}

{The cross-validation folds are stratified by participant (Section~\ref{sec:training}), so repetitions of each participant appear in both the training and test folds, and the evaluation does not establish how the approaches generalize to participants unseen during training. Performance under a subject-independent protocol such as leave-one-subject-out is typically lower~\citep{spilzBoostingAutomaticExercise2025}. We regard such an evaluation as a necessary step before deployment and leave it to future work.}

{The comparison also fixes the network architecture. Holding it constant isolates the effect of the training target (Section~\ref{sec:training}) but leaves open whether the advantage of the distribution-valued target transfers to other architecture families, such as recurrent, temporal--convolutional, transformer, or multi-stream networks. Establishing this remains for future work.}

The remaining limitations concern the empirical basis. The four datasets come from our own measurement studies with two fixed IMU setups and without external validation, so the transfer to other exercises, populations, or sensor configurations is untested.

\section{Conclusions}
\label{sec:conclusion}

This work introduced automatically generated label distributions, which extend a rule-based labeling procedure with simulated rater variability and yield a class distribution per repetition without a large human rater pool. On this basis we trained a network{, following established deep label distribution learning,} to reproduce the full distribution with a Kullback--Leibler objective, the ambiguity approach{. We} compared it against a one-hot cross-entropy baseline on four movement assessment datasets.
The ambiguity approach achieved at least the same performance as the baseline in classifying the most relevant class, and it detected ambiguous repetitions and their two relevant classes more reliably across the informative threshold range. Representing the label distribution in the training target therefore adds information about the ambiguity of a repetition at no cost to its classification. {Three directions remain open. The first is the treatment of repetitions in which more than two classes compete. The second is the extension of the rater model beyond threshold variability toward properties such as rater experience and observation perspective. The third is the calibration of the simulated rater variability and the entropy threshold against rater perception, which requires the collection of a dataset in which raters mark each repetition as ambiguous or unambiguous directly.}

\vspace{6pt} 

\authorcontributions{Conceptualization, A.S., H.O. and M.M.; methodology, A.S., H.O. and M.M.; software, A.S. and M.M.; validation, A.S. and M.M.; formal analysis, A.S. and M.M.; investigation, A.S. and M.M.; resources, M.M.; data curation, A.S.; writing---original draft preparation, A.S.; \mbox{writing---review} and editing, H.O. and M.M.; visualization, A.S.; supervision, M.M.; project administration, M.M.; funding acquisition, M.M. All authors have read and agreed to the published version of the manuscript.}

\funding{This research was funded by the Carl Zeiss Foundation (Carl-Zeiss-Stiftung) as part of the OrthoKI project, grant number P2022-07-009.}

\institutionalreview{The study was conducted in accordance with the Declaration of Helsinki and approved by the Ethics Committee of Ulm University of Applied Sciences (protocol code 2021-01, approved 17 May 2021, and protocol code 2024-01, approved 10 April 2024).}

\informedconsent{Written informed consent was obtained from all subjects involved in the study. A written consent form was distributed to all participants prior to the measurements and signed by each participant.}

\dataavailability{A subset of the analyzed data is publicly available via Zenodo at \url{https://zenodo.org/records/15729056}, accessed on 1 June 2026.
The remaining data are not publicly accessible but can be made available by the corresponding author upon reasonable request and subject to institutional data sharing agreements.}

\acknowledgments{During the preparation of this manuscript, the authors used Claude Opus 4.7 (Anthropic)
for language editing and to support the structuring of the text. The authors have reviewed and edited the output and take full responsibility for the content of this publication.}

\conflictsofinterest{The authors declare no conflicts of interest. The funders had no role in the design of the study; in the collection, analyses, or interpretation of data; in the writing of the manuscript; or in the decision to publish the results.}

\appendixtitles{yes} 
\appendixstart
\appendix
\section[\appendixname~\thesection]{Outcome Trees for RD, DS, and HS}
\label{app:outcome-trees}

\begin{figure}[H]

\includegraphics[width=0.85\textwidth]{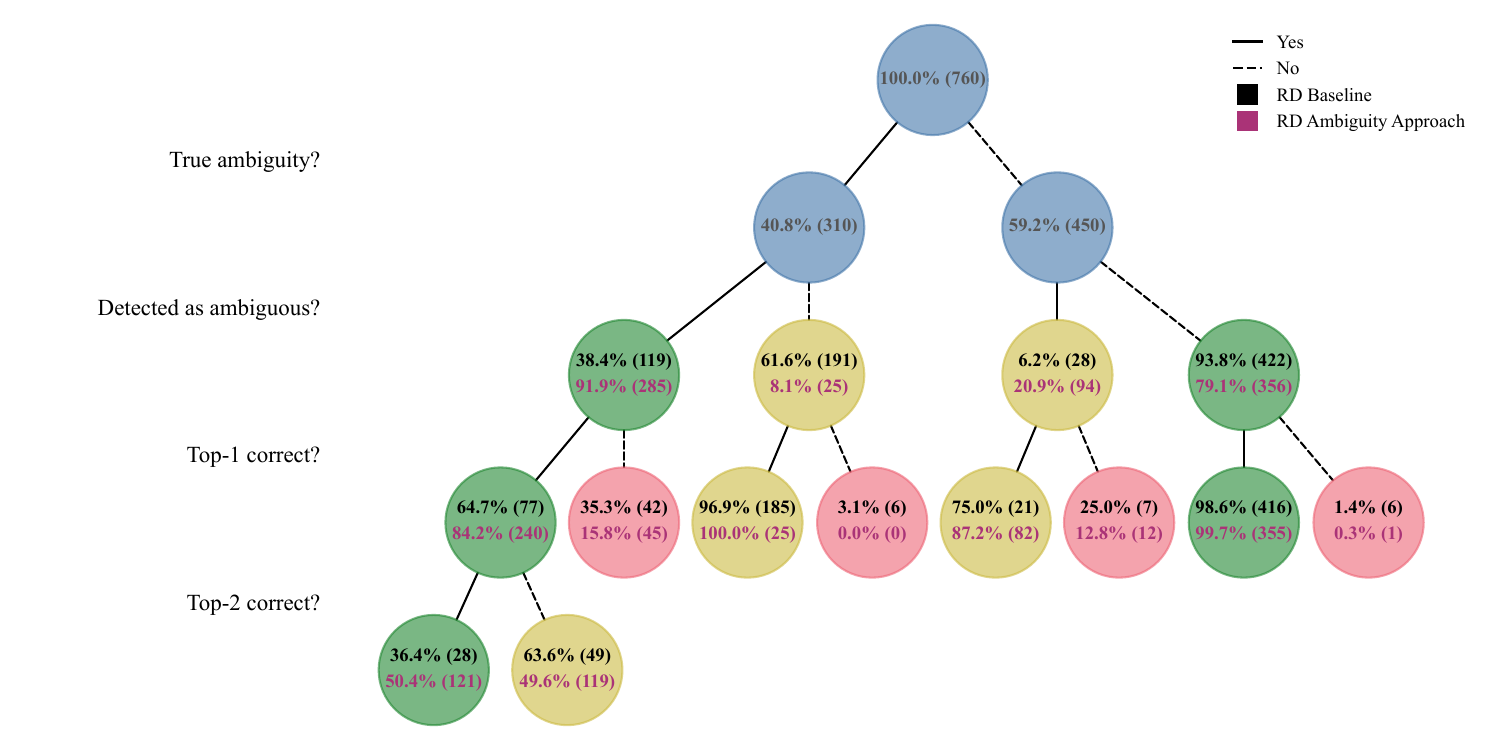}
\caption{Outcome tree for RD at the {exemplary} entropy threshold $\tau = 0.47$~bit. Construction and color coding follow Figure~\ref{fig:tree-rgs} (Section~\ref{sec:results-tree}).\label{fig:tree-rd}}
\end{figure}

\begin{figure}[H]

\includegraphics[width=0.85\textwidth]{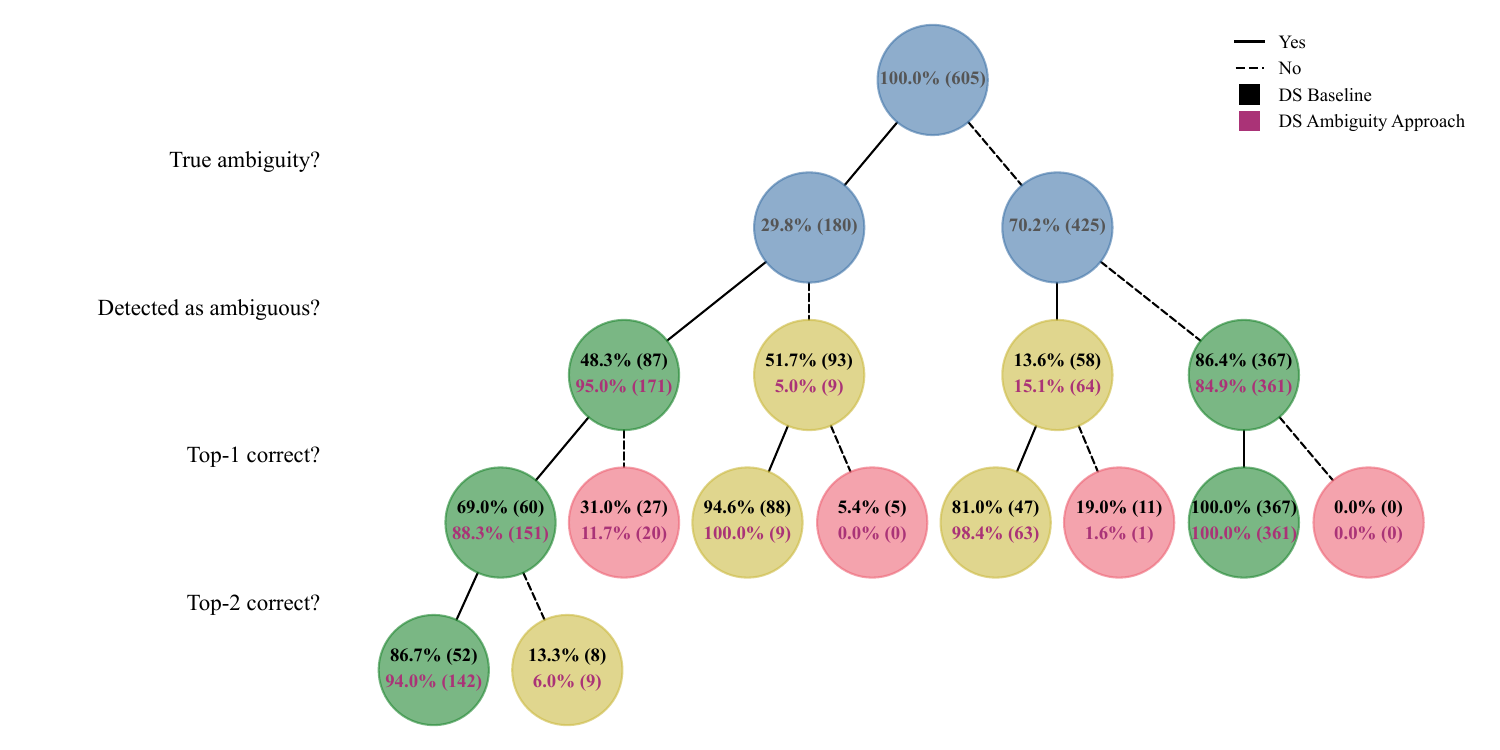}
\caption{Outcome tree for DS at the {exemplary} entropy threshold $\tau = 0.47$~bit. Construction and color coding follow Figure~\ref{fig:tree-rgs} (Section~\ref{sec:results-tree}).\label{fig:tree-ds}}
\end{figure}

\begin{figure}[H]

\includegraphics[width=0.85\textwidth]{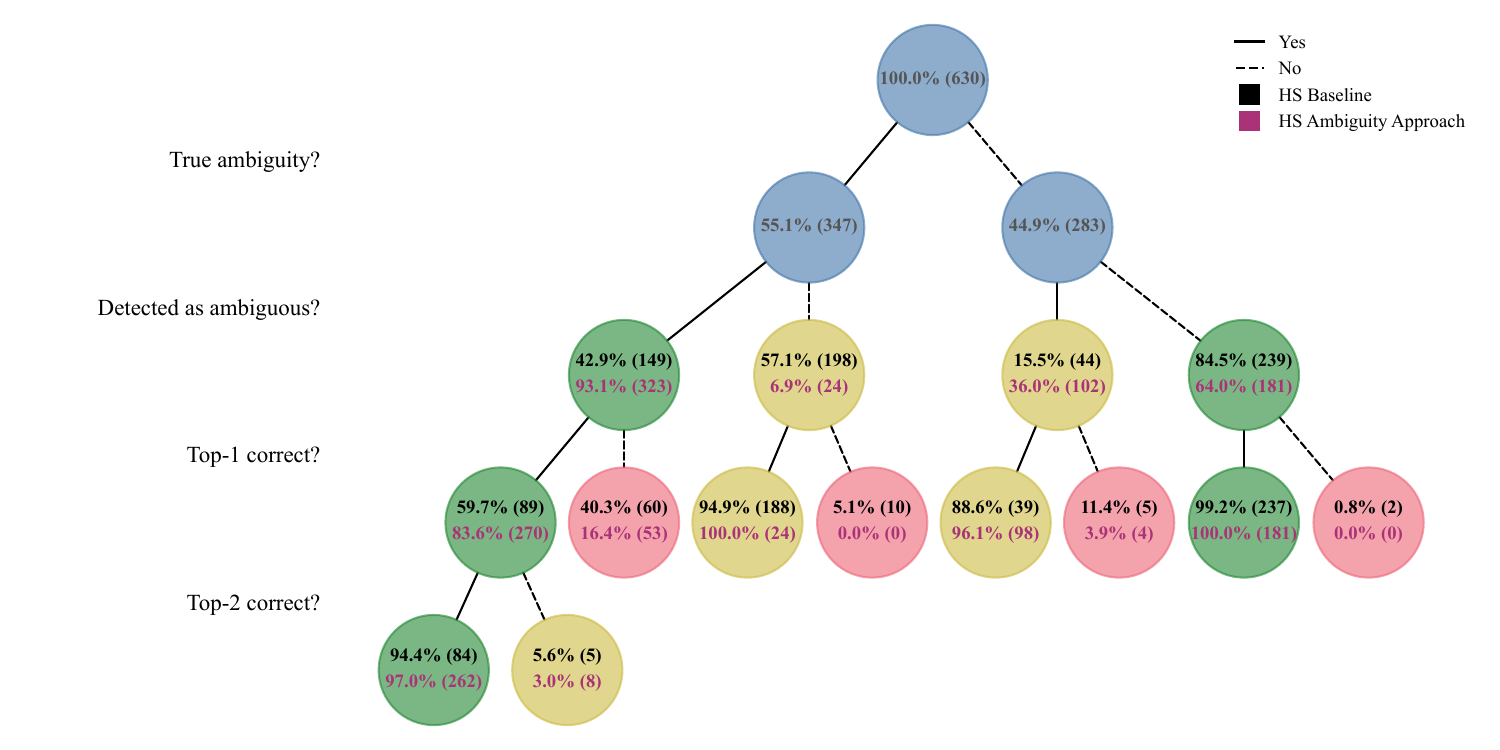}
\caption{Outcome tree for HS at the {exemplary} entropy threshold $\tau = 0.47$~bit. Construction and color coding follow Figure~\ref{fig:tree-rgs} (Section~\ref{sec:results-tree}).\label{fig:tree-hs}}
\end{figure}

This appendix provides the outcome trees for RD, DS, and HS (Figures~\ref{fig:tree-rd}--\ref{fig:tree-hs}) at the {exemplary} entropy threshold $\tau = 0.47$~bit, complementing the RGS tree in Figure~\ref{fig:tree-rgs}. Each tree follows the same construction and color coding described in Section~\ref{sec:results-tree}.
\vspace{-4pt}

\vspace{-16pt}

\section[\appendixname~\thesection]{Results of the Sensitivity Analysis}
\label{app:sensitivity}

{This appendix provides the table and figures of the sensitivity analysis defined in Section~\ref{sec:sensitivity} and reported in Section~\ref{sec:results-sensitivity}.}

\vspace{-4pt}
\begin{figure}[H]

\includegraphics[width=0.85\textwidth]{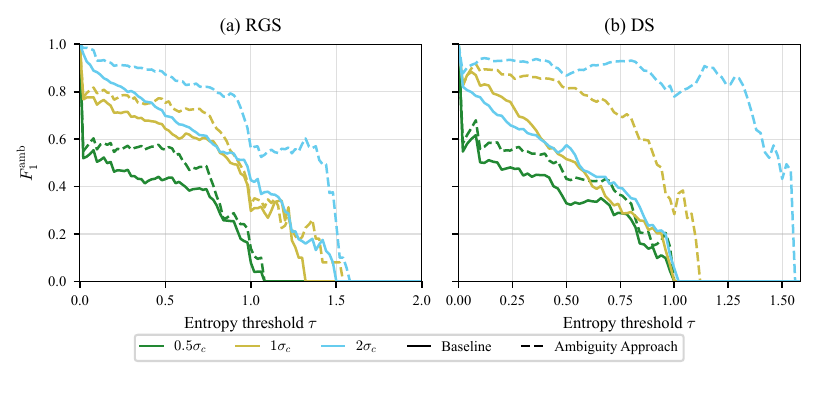}
\caption{{Ambiguity detection performance $F_1^{\mathrm{amb}}$ as a function of the entropy threshold $\tau$ in the five-fold cross-validation for the three variants of the sensitivity analysis (Section~\ref{sec:sensitivity}), with one panel per dataset. Solid lines show the baseline, dashed lines show the ambiguity approach, and colors distinguish variants $0.5\,\sigma_c$, $1\,\sigma_c$, and $2\,\sigma_c$. Lines give the mean over the five folds. The threshold extends to the maximum attainable entropy of each dataset, $\log_2 4 = 2$~bit for RGS and $\log_2 3 \approx 1.58$~bit for DS.}\label{fig:sensitivity-ambdet}}
\end{figure}

\begin{table}[H]

\caption{{Classification performance of the sensitivity analysis in the five-fold cross-validation. $F_1^{\mathrm{cls}}$ and class-wise $F_1$-scores (mean $\pm$ standard deviation over the five folds) for the baseline and the ambiguity approach on RGS and DS under the three variants, $0.5\,\sigma_c$, $1\,\sigma_c$, and $2\,\sigma_c$ (Section~\ref{sec:sensitivity}). The $1\,\sigma_c$ rows repeat the main configuration from Table~\ref{tab:classification-tratr}. The class-wise columns $F_{1,1}^{\mathrm{cls}}$ to $F_{1,4}^{\mathrm{cls}}$ follow the class order per exercise, RGS (CE, IKF, IHA, and SH) and DS (FMS~1, 2, and 3). DS has only three classes, so $F_{1,4}^{\mathrm{cls}}$ is empty.}\label{tab:sensitivity-classification}}

	\isPreprints{}{\begin{adjustwidth}{-\extralength}{0cm}}

		\begin{tabularx}{\widetablelength}{LlCCCCC}
\toprule
\textbf{Exercise} & \textbf{Setup} & \boldmath{$F_1^{\mathrm{cls}}$} & \boldmath{$F_{1,1}^{\mathrm{cls}}$} & \boldmath{$F_{1,2}^{\mathrm{cls}}$} & \boldmath{$F_{1,3}^{\mathrm{cls}}$} & \boldmath{$F_{1,4}^{\mathrm{cls}}$} \\
\midrule
\multirow[c]{6}{*}{RGS} & Baseline ($0.5\,\sigma_c$) & 0.89 $\pm$ 0.02 & 0.88 $\pm$ 0.03 & 0.83 $\pm$ 0.04 & 0.94 $\pm$ 0.02 & 0.90 $\pm$ 0.04 \\
 & Ambiguity Approach ($0.5\,\sigma_c$) & 0.90 $\pm$ 0.02 & 0.89 $\pm$ 0.03 & 0.86 $\pm$ {0.04} & 0.96 $\pm$ {0.02} & 0.91 $\pm$ 0.03 \\
 & Baseline ($1\,\sigma_c$) & 0.89 $\pm$ {0.02} & 0.89 $\pm$ {0.06} & 0.82 $\pm$ 0.02 & 0.95 $\pm$ 0.03 & 0.89 $\pm$ 0.01 \\
 & Ambiguity Approach ($1\,\sigma_c$) & 0.90 $\pm$ 0.02 & 0.91 $\pm$ {0.05} & 0.84 $\pm$ 0.02 & 0.96 $\pm$ 0.02 & 0.89 $\pm$ 0.01 \\
 & Baseline ($2\,\sigma_c$) & 0.88 $\pm$ 0.03 & 0.85 $\pm$ 0.06 & 0.84 $\pm$ {0.05} & 0.94 $\pm$ {0.05} & 0.90 $\pm$ {0.02} \\
 & Ambiguity Approach ($2\,\sigma_c$) & 0.91 $\pm$ 0.02 & 0.90 $\pm$ {0.04} & 0.88 $\pm$ 0.04 & 0.95 $\pm$ {0.03} & 0.90 $\pm$ {0.03} \\
\midrule
\multirow[c]{6}{*}{DS} & Baseline ($0.5\,\sigma_c$) & 0.93 $\pm$ 0.02 & 0.94 $\pm$ 0.02 & 0.85 $\pm$ {0.04} & 1.00 $\pm$ 0.00 &  \\
 & Ambiguity Approach ($0.5\,\sigma_c$) & 0.95 $\pm$ 0.01 & 0.96 $\pm$ 0.01 & 0.89 $\pm$ 0.03 & 1.00 $\pm$ 0.00 &  \\
 & Baseline ($1\,\sigma_c$) & 0.93 $\pm$ 0.01 & 0.95 $\pm$ 0.01 & 0.86 $\pm$ {0.03} & 1.00 $\pm$ 0.00 &  \\
 & Ambiguity Approach ($1\,\sigma_c$) & 0.97 $\pm$ {0.02} & 0.98 $\pm$ 0.01 & 0.92 $\pm$ {0.04} & 1.00 $\pm$ 0.00 &  \\
 & Baseline ($2\,\sigma_c$) & 0.94 $\pm$ 0.02 & 0.95 $\pm$ 0.02 & 0.86 $\pm$ 0.05 & 1.00 $\pm$ 0.00 &  \\
 & Ambiguity Approach ($2\,\sigma_c$) & 0.95 $\pm$ {0.04} & 0.97 $\pm$ 0.02 & 0.87 $\pm$ {0.09} & 1.00 $\pm$ 0.00 &  \\

\bottomrule
\end{tabularx}
\isPreprints{}{\end{adjustwidth}}
\end{table}

\vspace{-16pt}

\begin{figure}[H]

\includegraphics[width=0.85\textwidth]{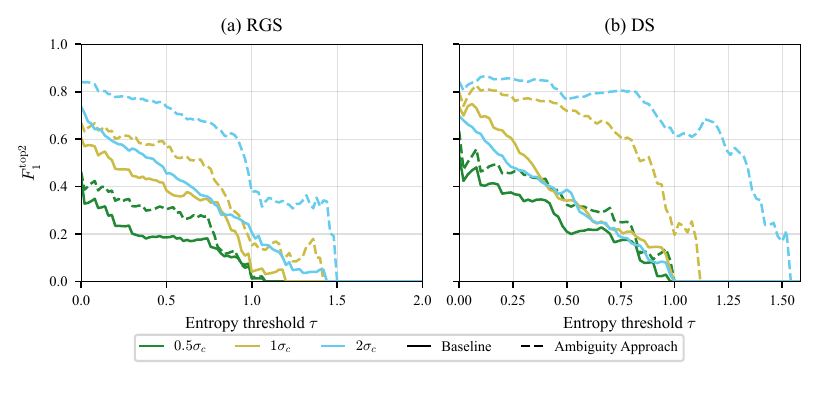}
\caption{{Combined ambiguity detection and top-2 performance $F_1^{\mathrm{top2}}$ as a function of the entropy threshold $\tau$ in the five-fold cross-validation for the three variants of the sensitivity analysis \mbox{(Section~\ref{sec:sensitivity})}, with one panel per dataset. Solid lines show the baseline, dashed lines show the ambiguity approach, and colors distinguish variants $0.5\,\sigma_c$, $1\,\sigma_c$, and $2\,\sigma_c$. Lines give the mean over the five folds. The threshold extends to the maximum attainable entropy of each dataset, $\log_2 4 = 2$~bit for RGS and $\log_2 3 \approx 1.58$~bit for DS.}\label{fig:sensitivity-top2}}
\end{figure}

\isPreprints{}{
\begin{adjustwidth}{-\extralength}{0cm}
}

\reftitle{References}



\isPreprints{}{\PublishersNote{}}
\isPreprints{}{%
\end{adjustwidth}
}

\begin{thebibliography}{999}

\bibitem[Ashari et~al.(2016)Ashari, Hamid, Hussain, and Hill]{ashari_effectiveness_2016}
Ashari, A.; Hamid, T.A.; Hussain, M.R.; Hill, K.D.
\newblock Effectiveness of {Individualized} {Home}-{Based} {Exercise} on {Turning} and {Balance} {Performance} {Among} {Adults} {Older} than 50 yrs.
\newblock {\em Am. J. Phys. Med. Rehabil.} {\bf 2016}, {\em 95},~355--365.
\newblock {\url{https://doi.org/10.1097/phm.0000000000000388}}.

\bibitem[Latham et~al.(2014)Latham, Harris, Bean, Heeren, Goodyear, Zawacki, Heislein, Mustafa, Pardasaney, Giorgetti, Holt, Goehring, and Jette]{latham_effect_2014}
Latham, N.K.; Harris, B.A.; Bean, J.F.; Heeren, T.; Goodyear, C.; Zawacki, S.; Heislein, D.M.; Mustafa, J.; Pardasaney, P.; Giorgetti, M.;  et~al.
\newblock Effect of a {Home}-{Based} {Exercise} {Program} on {Functional} {Recovery} {Following} {Rehabilitation} {After} {Hip} {Fracture}: {A} {Randomized} {Clinical} {Trial}.
\newblock {\em JAMA} {\bf 2014}, {\em 311},~700--708.
\newblock {\url{https://doi.org/10.1001/jama.2014.469}}.

\bibitem[Gelaw et~al.(2020)Gelaw, Janakiraman, Gebremeskel, and Ravichandran]{gelaw_effectiveness_2020}
Gelaw, A.Y.; Janakiraman, B.; Gebremeskel, B.F.; Ravichandran, H.
\newblock Effectiveness of {Home}-based rehabilitation in improving physical function of persons with {Stroke} and other physical disability: {A} systematic review of randomized controlled trials.
\newblock \linebreak  {\em J. Stroke Cerebrovasc. Dis. Off. J. Natl. Stroke Assoc.} {\bf 2020}, {\em 29},~104800.
\newblock {\url{https://doi.org/10.1016/j.jstrokecerebrovasdis.2020.104800}}.

\bibitem[Flynn et~al.(2019)Flynn, Allen, Dennis, Canning, and Preston]{flynn_home-based_2019}
Flynn, A.; Allen, N.E.; Dennis, S.; Canning, C.G.; Preston, E.
\newblock Home-based prescribed exercise improves balance-related activities in people with {Parkinson}'s disease and has benefits similar to centre-based exercise: A systematic review.
\newblock {\em J. Physiother.} {\bf 2019}, {\em 65},~189--199.
\newblock {\url{https://doi.org/10.1016/j.jphys.2019.08.003}}.

\bibitem[Argent et~al.(2018)Argent, Daly, and Caulfield]{argent_patient_2018}
Argent, R.; Daly, A.; Caulfield, B.
\newblock Patient {Involvement} {with} {Home}-{Based} {Exercise} {Programs}: {Can} {Connected} {Health} {Interventions} {Influence} {Adherence}?
\newblock {\em JMIR mHealth uHealth} {\bf 2018}, {\em 6},~e47.
\newblock {\url{https://doi.org/10.2196/mhealth.8518}}.

\bibitem[Faber et~al.(2015)Faber, Andersen, Sevel, Thorborg, Bandholm, and Rathleff]{faber_majority_2015}
\textls[-25]{Faber, M.; Andersen, M.H.; Sevel, C.; Thorborg, K.; Bandholm, T.; Rathleff, M.
\newblock The majority are not performing home-exercises correctly two weeks after their initial instruction---An assessor-blinded study.
\newblock {\em PeerJ} {\bf 2015}, {\em 3},~e1102.
\newblock {\url{https://doi.org/10.7717/peerj.1102}}.}

\bibitem[Lang et~al.(2022)Lang, McLelland, MacDonald, and Hamilton]{lang_digital_2022}
\textls[-15]{Lang, S.; McLelland, C.; MacDonald, D.; Hamilton, D.F.
\newblock Do digital interventions increase adherence to home exercise rehabilitation? {A} systematic review of randomised controlled trials.
\newblock {\em Arch. Physiother.} {\bf 2022}, {\em 12},~24.
\newblock {\url{https://doi.org/10.1186/s40945-022-00148-z}}.}

\bibitem[Swain et~al.(2023)Swain, McNarry, Runacres, and Mackintosh]{swainRoleMultiSensorMeasurement2023}
Swain, T.A.; McNarry, M.A.; Runacres, A.W.H.; Mackintosh, K.A.
\newblock The {Role} of {Multi}-{Sensor} {Measurement} in the {Assessment} of {Movement} {Quality}: {A} {Systematic} {Review}.
\newblock {\em Sport. Med.} {\bf 2023}, {\em 53},~2477--2504.
\newblock {\url{https://doi.org/10.1007/s40279-023-01905-1}}.

\bibitem[Lee et~al.(2020)Lee, Joo, Lee, and Chee]{leeAutomaticClassificationSquat2020}
Lee, J.; Joo, H.; Lee, J.; Chee, Y.
\newblock Automatic {Classification} of {Squat} {Posture} {Using} {Inertial} {Sensors}: {Deep} {Learning} {Approach}.
\newblock {\em Sensors} {\bf 2020}, {\em 20},~361.
\newblock {\url{https://doi.org/10.3390/s20020361}}.

\bibitem[Cook et~al.(2014{\natexlab{a}})Cook, Burton, Hoogenboom, and Voight]{cook_functional_2014}
Cook, G.; Burton, L.; Hoogenboom, B.J.; Voight, M.
\newblock Functional movement screening: The use of fundamental movements as an assessment of function---Part 1.
\newblock {\em Int. J. Sport. Phys. Ther.} {\bf 2014}, {\em 9},~396--409.

\bibitem[Cook et~al.(2014{\natexlab{b}})Cook, Burton, Hoogenboom, and Voight]{cook_functional_2014-1}
Cook, G.; Burton, L.; Hoogenboom, B.J.; Voight, M.
\newblock Functional movement screening: The use of fundamental movements as an assessment of function---Part 2.
\newblock {\em Int. J. Sport. Phys. Ther.} {\bf 2014}, {\em 9},~549--563.

\bibitem[Shultz et~al.(2013)Shultz, Anderson, Matheson, Marcello, and Besier]{shultz_test-retest_2013}
Shultz, R.; Anderson, S.C.; Matheson, G.O.; Marcello, B.; Besier, T.
\newblock Test-{Retest} and {Interrater} {Reliability} of the {Functional} {Movement} {Screen}.
\newblock {\em J. Athl. Train.} {\bf 2013}, {\em 48},~331--336.
\newblock {\url{https://doi.org/10.4085/1062-6050-48.2.11}}.

\bibitem[Kottner et~al.(2009)Kottner, Raeder, Halfens, and Dassen]{kottner_systematic_2009}
Kottner, J.; Raeder, K.; Halfens, R.; Dassen, T.
\newblock A systematic review of interrater reliability of pressure ulcer classification systems.
\newblock {\em J. Clin. Nurs.} {\bf 2009}, {\em 18},~315--336.
\newblock {\url{https://doi.org/10.1111/j.1365-2702.2008.02569.x}}.

\bibitem[Dalton et~al.(2000)Dalton, Pinder, Elston, Ellis, Page, Dupont, and Blamey]{dalton_histologic_2000}
\textls[-25]{Dalton, L.W.; Pinder, S.E.; Elston, C.E.; Ellis, I.O.; Page, D.L.; Dupont, W.D.; Blamey, R.W.
\newblock Histologic Grading of Breast Cancer: Linkage of Patient Outcome with Level of Pathologist Agreement.
\newblock {\em Mod. Pathol.} {\bf 2000}, {\em 13},~730--735.
\newblock {\url{https://doi.org/10.1038/modpathol.3880126}}.}

\bibitem[Geng(2016)]{gengLabelDistributionLearning2016}
\textls[-25]{Geng, X.
\newblock Label {Distribution} {Learning}.
\newblock {\em IEEE Trans. Knowl. Data Eng.} {\bf 2016}, {\em 28},~1734--1748.
\newblock {\url{https://doi.org/10.1109/TKDE.2016.2545658}}.}

\bibitem[Gao et~al.(2017)Gao, Xing, Xie, Wu, and Geng]{gaoDeepLabelDistribution2017a}
Gao, B.B.; Xing, C.; Xie, C.W.; Wu, J.; Geng, X.
\newblock Deep {Label} {Distribution} {Learning} {with} {Label} {Ambiguity}.
\newblock {\em IEEE Trans. Image Process.} {\bf 2017}, {\em 26},~2825--2838.
\newblock {\url{https://doi.org/10.1109/TIP.2017.2689998}}.

\bibitem[Gao et~al.(2020)Gao, Zhang, and Geng]{gaoLabelEnhancementLabel2020}
Gao, Y.; Zhang, Y.; Geng, X.
\newblock Label {Enhancement} for {Label} {Distribution} {Learning} via {Prior} {Knowledge}.
\newblock In Proceedings of the {Twenty}-{Ninth} {International} {Joint} {Conference} on {Artificial} {Intelligence} ({IJCAI}-20),  Yokohama, Japan, 11--17 July 2020; \mbox{pp. 3223--3229}.
\newblock {\url{https://doi.org/10.24963/ijcai.2020/446}}.

\bibitem[Lienen and Hüllermeier(2024)]{lienenMitigatingLabelNoise2024}
Lienen, J.; Hüllermeier, E.
\newblock Mitigating label noise through data ambiguation.
\newblock In \emph{Proceedings of the {Thirty}-{Eighth} {AAAI} {Conference} on {Artificial} {Intelligence} and {Thirty}-{Sixth} {Conference} on {Innovative} {Applications} of {Artificial} {Intelligence} and {Fourteenth} {Symposium} on {Educational} {Advances} in {Artificial} {Intelligence}}; AAAI Press:  Washington, DC, USA, 2024; Volume~38, pp. 13799--13807.
\linebreak  \newblock {\url{https://doi.org/10.1609/aaai.v38i12.29286}}.

\bibitem[Li et~al.(2023)Li, Sun, and Li]{liConfusionMatrixLearning2023}
Li, J.; Sun, H.; Li, J.
\newblock Beyond confusion matrix: Learning from multiple annotators with awareness of instance features.
\newblock {\em Mach. Learn.} {\bf 2023}, {\em 112},~1053--1075.
\newblock {\url{https://doi.org/10.1007/s10994-022-06211-x}}.

\bibitem[Chen et~al.(2017)Chen, Zhang, Dong, Le, and Rao]{chenUsingRankingCNNAge2017}
Chen, S.; Zhang, C.; Dong, M.; Le, J.; Rao, M.
\newblock Using {Ranking}-{CNN} for {Age} {Estimation}.
\newblock In Proceedings of the {IEEE} {Conference} on {Computer} {Vision} and {Pattern} {Recognition} ({CVPR}),  Honolulu, HI, USA, 21--26 July 2017; pp. 5183--5192.

\bibitem[Lin et~al.(2023)Lin, Huang, Ruan, Yang, Chen, Zheng, and Feng]{linAutomaticEvaluationFunctional2023}
\textls[-15]{Lin, X.; Huang, T.; Ruan, Z.; Yang, X.; Chen, Z.; Zheng, G.; Feng, C.
\newblock Automatic {Evaluation} of {Functional} {Movement} {Screening} {Based} on {Attention} {Mechanism} and {Score} {Distribution} {Prediction}.
\newblock {\em Mathematics} {\bf 2023}, {\em 11},~4936.
\newblock {\url{https://doi.org/10.3390/math11244936}}.}

\bibitem[Lin et~al.(2024{\natexlab{a}})Lin, Chen, Feng, Chen, Yang, and Cui]{linAutomaticEvaluationMethod2024}
Lin, X.; Chen, R.; Feng, C.; Chen, Z.; Yang, X.; Cui, H.
\newblock Automatic {Evaluation} {Method} for {Functional} {Movement} {Screening} {Based} on a {Dual}-{Stream} {Network} and {Feature} {Fusion}.
\newblock {\em Mathematics} {\bf 2024}, {\em 12},~1162.
\newblock {\url{https://doi.org/10.3390/math12081162}}.

\bibitem[Lin et~al.(2024{\natexlab{b}})Lin, Liu, Feng, Chen, Yang, and Cui]{linAutomaticEvaluationMethod2024a}
\textls[-15]{Lin, X.; Liu, Y.; Feng, C.; Chen, Z.; Yang, X.; Cui, H.
\newblock Automatic {Evaluation} {Method} for {Functional} {Movement} {Screening} {Based} on {Multi}-{Scale} {Lightweight} {3D} {Convolution} and an {Encoder}–{Decoder}.
\newblock {\em Electronics} {\bf 2024}, {\em 13},~1813.
\newblock {\url{https://doi.org/10.3390/electronics13101813}}.}

\bibitem[Spilz et~al.(2025)Spilz, Oppel, and Munz]{spilzBoostingAutomaticExercise2025}
Spilz, A.; Oppel, H.; Munz, M.
\newblock Boosting {Automatic} {Exercise} {Evaluation} {Through} {Musculoskeletal} {Simulation}-{Based} {IMU} {Data} {Augmentation}. \emph{arXiv} \textbf{2025},
\newblock arXiv:2505.24415. {\url{https://doi.org/10.48550/arXiv.2505.24415}}.

\bibitem[Spilz and Munz(2023)]{spilz_automatic_2023}
Spilz, A.; Munz, M.
\newblock Automatic {Assessment} of {Functional} {Movement} {Screening} {Exercises} with {Deep} {Learning} {Architectures}.
\newblock {\em Sensors} {\bf 2023}, {\em 23},~5.
\newblock {\url{https://doi.org/10.3390/s23010005}}.

\bibitem[Spilz et~al.(2026)Spilz, Oppel, Werner, Stucke-Straub, Capanni, and Munz]{spilz_gaitex_2025}
Spilz, A.; Oppel, H.; Werner, J.; Stucke-Straub, K.; Capanni, F.; Munz, M.
\newblock {GAITEX}: {Human} motion dataset of impaired gait and rehabilitation exercises using inertial and optical sensors.
\newblock {\em Sci. Data} {\bf 2026}, {\em 13},~11.
\newblock {\url{https://doi.org/10.1038/s41597-025-06439-x}}.

\bibitem[Roetenberg et~al.(2009)Roetenberg, Luinge, and Slycke]{roetenberg}
Roetenberg, D.; Luinge, H.; Slycke, P.
\newblock Xsens MVN: Full 6DOF human motion tracking using miniature inertial sensors.
\newblock {\em Xsens Motion Technol. BV Tech. Rep.} {\bf 2009}, {\em 3}.
\newblock \url{https://api.semanticscholar.org/CorpusID:16142980}.


\bibitem[Delp et~al.(2007)Delp, Anderson, Arnold, Loan, Habib, John, Guendelman, and Thelen]{delp_opensim_2007}
Delp, S.L.; Anderson, F.C.; Arnold, A.S.; Loan, P.; Habib, A.; John, C.T.; Guendelman, E.; Thelen, D.G.
\newblock {OpenSim}: Open-source software to create and analyze dynamic simulations of movement.
\newblock {\em IEEE Trans. Biomed. Eng.} {\bf 2007}, {\em 54},~1940--1950.
\newblock {\url{https://doi.org/10.1109/TBME.2007.901024}}.

\bibitem[Shoemake(1985)]{shoemake_animating_1985}
Shoemake, K.
\newblock Animating rotation with quaternion curves.
\newblock {\em SIGGRAPH Comput. Graph.} {\bf 1985}, {\em 19},~245--254.
\newblock {\url{https://doi.org/10.1145/325165.325242}}.

\end{thebibliography}
\end{document}